\documentclass{article} 
\usepackage{collas2022_conference,times}

\collasfinalcopy

\usepackage{hyperref}
\hypersetup{
	colorlinks=true,
	linkcolor=red,
	filecolor=magenta,
	urlcolor=blue,
	citecolor=purple,
	pdftitle={Overleaf Example},
	pdfpagemode=FullScreen,
	}

\usepackage{amsmath}
\usepackage{amsfonts}
\usepackage{amssymb}
\usepackage{amsthm}
\usepackage{tabularx}
\usepackage{booktabs}
\usepackage{graphicx}
\usepackage{hyperref}
\usepackage{bm}
\usepackage{multirow}
\usepackage{bbm}
\usepackage{enumitem}
\usepackage{adjustbox}

\newcolumntype{P}[1]{>{\raggedright\arraybackslash}p{#1}}

\usepackage{tikz}
\usepackage{caption}
\usepackage{subcaption}
\usepackage{fancyhdr}

\usepackage{algorithm2e}

\usepackage{amssymb}
\usepackage{pifont}

\theoremstyle{definition}

\newcommand{\dtrain}{\mathcal{D}_{\text{train}}}

\newcommand{\dvalid}{\mathcal{D}_{\text{valid}}}
\newcommand{\dsupp}{\mathcal{D}^{(k)}_{\text{valid}}}

\newcommand{\dtest}{\mathcal{D}_{\text{test}}}
\newcommand{\daug}{\tilde{\mathcal{D}}^{(k)}_{\text{aug}}}
\newcommand{\ytrain}{\mathcal{Y}_{\text{train}}}
\newcommand{\yvalid}{\mathcal{Y}_{\text{valid}}}
\newcommand{\ytest}{\mathcal{Y}_{\text{test}}}

\newcommand{\sigm}{\text{sigm}}

\newcommand{\dfm}{\emph{dfm} }
\newcommand{\gfm}{\emph{gfm} }

\title{Overcoming challenges in leveraging GANs for few-shot data augmentation}

\author{Christopher Beckham$^{1,2,3}$, Issam Laradji$^{1}$, Pau Rodriguez$^{1}$, David Vazquez$^{1}$, Derek Nowrouzezahrai$^{2,4}$, \\ \textbf{Christopher Pal}$^{1,2,3,\dagger}$ \\
$^{1}$ServiceNow Research, $^{2}$Mila, $^{3}$Polytechnique Montreal, $^{4}$McGill University, $^{\dagger}$Canada CIFAR AI Chair
}

\begin{document}

\maketitle
\begin{abstract}
In this paper, we explore the use of GAN-based few-shot data augmentation as a method to improve few-shot classification performance. We perform an exploration into how a GAN can be fine-tuned for such a task (one of which is in a \emph{class-incremental} manner), as well as a rigorous empirical investigation into how well these models can perform to improve few-shot classification. We identify issues related to the difficulty of training and applying such generative models under a purely supervised regime with very few examples, as well as issues regarding the evaluation protocols of existing works. We also find that in this regime, classification accuracy is highly sensitive to how the classes of the dataset are randomly split. To address difficulties in applying these generative models under the few-shot regime, we propose a simple and pragmatic semi-supervised fine-tuning approach, and demonstrate gains in FID and precision-recall metrics as well as classification performance.
\end{abstract}
\section{Introduction}


In the past decade, deep learning has demonstrated immense success in many tasks and data modalities~\citep{krizhevsky2012imagenet, lecun2015deep, he2016deep}. However, the issue of deep neural networks being highly data inefficient remains a major problem \citep{krizhevsky2012imagenet, srivastava2014dropout, mixup, tan2018survey}, accentuated by the fact that for some domains, obtaining sufficient labelled data is laborious and expensive. This  limits real-world applicability of deep neural networks, so one active area of research is in figuring out how to make these models transfer or learn better on under-represented datasets. Few-shot learning is one of the subfields that addresses this, where the goal is to facilitate adaptation to novel tasks with very few examples.

Few-shot learning as a field is rather broad \citep{wang2020generalizing}, and many techniques can be seen as incorporating prior knowledge into either the data, the model/architecture, or the algorithm in order to make the learning process more sample efficient. For example, this may be in the form of multitask learning \citep{caruana1997multitask, zhang2017survey} where the learning of a few-shot task can be augmented by learning other relevant tasks with more data, or in embedding learning \citep{bertinetto2016learning, sung2018learning, vinyals2016matching} where a low-dimensional metric space is learned to be able to easily facilitate comparisons between unseen classes. In terms of algorithmic few-shot learning, meta-learning \citep{maml, ravi2016optimization}, i.e. \emph{learning how to learn}, can be used to learn more sample efficient optimisers. 

In this paper, we focus on the \emph{data} aspect of few-shot learning, which at its core leverages data augmentation to expand a small set of examples into a much larger set that will be sufficient to train deep neural networks. In particular, we explore data augmentation with \emph{generative adversarial networks} (GANs) \citep{jsgan}, where the goal is to generate samples that are indistinguishable from the ones comprising the few-shot task of interest. Concretely, our contributions are as follows:
\begin{itemize}
	\item We consider the entire few-shot data augmentation pipeline, where the goal is to pre-train a GAN on some set of source (training) classes, fine-tune it on very few examples of target (test) classes, and have it be able to generate new images from those classes to improve generalisation performance of a pre-trained classifier on the same target classes. Unlike previous works \citep{dagan, hong2020matchinggan, f2gan}, we specifically fine-tune our GAN on novel classes rather than try to generate from them in `zero shot' fashion.
	\item We rigorously evaluate few-shot data augmentation performance by considering different randomised splits of the dataset (\emph{dataset seeds}), where the splits are performed with respect to the \emph{classes}. We also perform these experiments over different fine-tuning strategies where different parts of both the generator and discriminator are frozen, with the adversarial game still being played between both networks. 
	\item We highlight the difficulties involved in few-shot data augmentation, and explain limitations with the empirical evaluation of existing work. To this end, we propose a more pragmatic strategy involving semi-supervised GAN fine-tuning, and demonstrate performance gains with respect to classification performance, as well as FID and precision/recall. As a side contribution, one of our proposed fine-tuning strategies enables the generator to learn new classes in a \emph{class-incremental} manner, without catastrophic forgetting of old ones. This may be of interest to those working in the intersection of generative modelling and \emph{generalised} few-shot learning.
\end{itemize}

\subsection{Few-shot data augmentation pipeline} \label{sec:problem_setup}

\begin{figure}[h]
	\centering
	\includegraphics[width=0.9\textwidth,page=1]{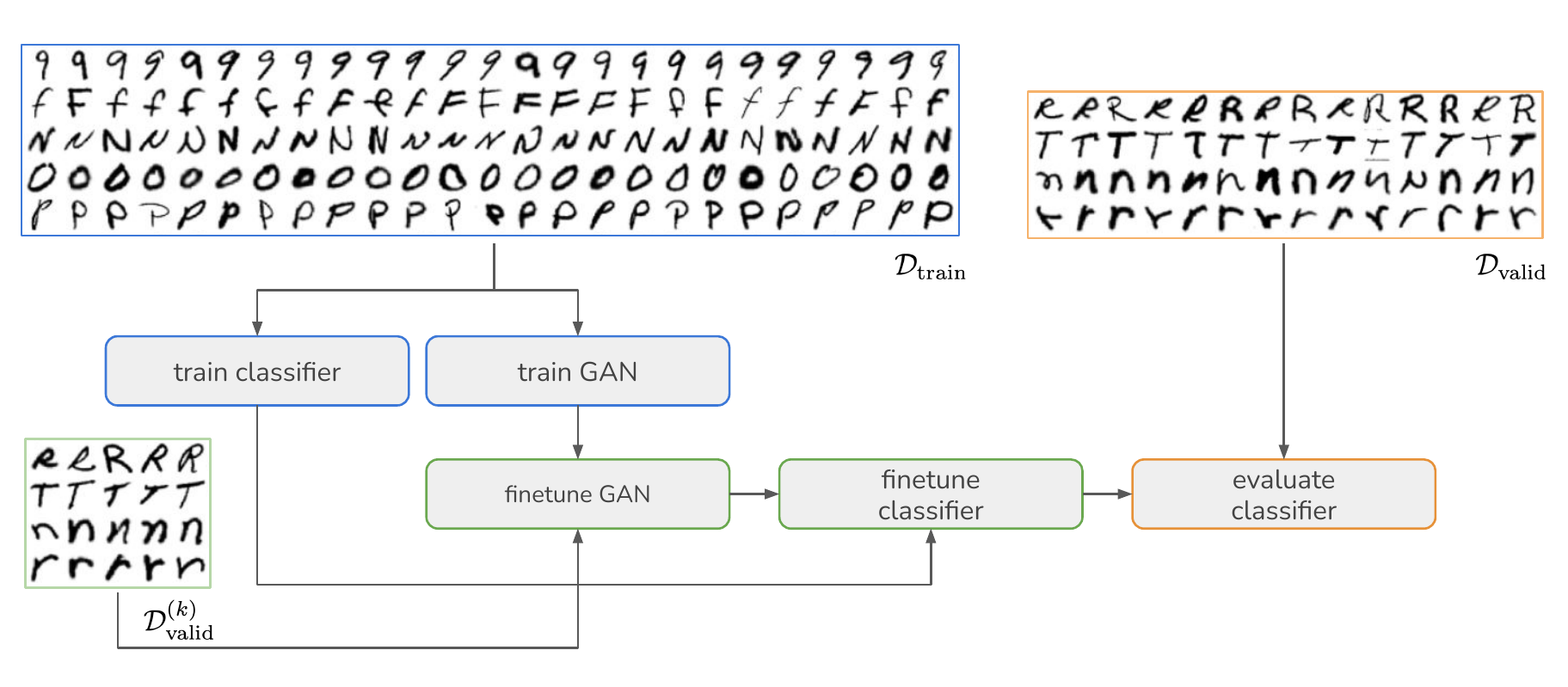}
	\caption{The few-shot data augmentation pipeline. In this illustration, the training set consists of five classes (abundantly represented) and the validation set consists of four classes. The support set is a subset of the validation set and only comprises five examples per class ($k=5$). Each stage is described in Section \ref{sec:proposed}.}
	\label{fig:protocol}
\end{figure}

Our few-shot data augmentation pipeline is illustrated in Figure \ref{fig:protocol}. First, we divide the data into a training set $\dtrain$, validation set $\dvalid$, and test set $\dtest$, where $\ytrain \cap \yvalid = \emptyset$ and $\yvalid = \ytest$. We can think of the training and validation (test) sets as comprising our \emph{source} and \emph{target} classes, respectively. We also assume that each source class is abundantly represented. Our goal is to train a conditional generative model that can generate plausible examples from the validation set $\dvalid$ given only a very small subset of that same set, which is called the `support set' $\dsupp$. The superscript $(k)$ denotes that there are only $k$ examples per target class that we can work with. Because this is an extremely difficult task, we pre-train the GAN on the source classes first (i.e. the training set), which we assume is abundantly represented. Our primary measure of success is whether the examples generated from our model are able to improve the generalisation performance of a classifier trained to predict the validation classes $\yvalid$.


Because the few-shot setting requires us to adapt our models to novel classes with very small amounts of data, the evaluation can be highly sensitive to how the dataset is split. Furthermore, it may also be sensitive to what classes comprise source classes (those in the training set) and what classes comprise novel classes (those in the validation or test set). Because of this, almost all quantitative results in this paper are averages over many randomised splits of the dataset in order to produce reliable estimates of uncertainty. (This is illustrated in Figure \ref{fig:appendix_data_splits}.)


\section{Proposed method} \label{sec:proposed}

We break down our training and evaluation pipeline into the following subsections:

\begin{enumerate}
\item Section \ref{sec:classifier_pretrain}: pre-train a classifier on $\dtrain$ which estimates $p(\bm{y} \in \ytrain |\bm{x})$. This classifier will eventually be fine-tuned to adapt to the new classes in $\dvalid$.
\item Sections \ref{sec:gan_training} and \ref{sec:gan_finetuning}: train a conditional GAN on $\dtrain$, where we learn a conditional generative model to be able to sample $\bm{x} \sim p_G(\bm{x}|\bm{y} \in \ytrain)$. Afterwards, we leverage some \emph{fine-tuning strategy} to fine-tune the GAN on $\dsupp$, which will allow sample generation for the novel classes in $\yvalid$.
\item Use the fine-tuned GAN to generate new examples for each \emph{target class}. These examples will be concatenated with the original support set $\dsupp$ to produce an `augmented' set which we denote $\daug$.
\item Section \ref{sec:exps_and_results}: Replace the pre-trained classifier's output probability distribution to be over the novel classes $p(\bm{y} \in \yvalid | \bm{x})$ and fine-tune it by using $\daug$ as the training set and perform model selection (hyperparameter tuning) on $\dvalid$. The best model is evaluated on $\dtest$, which is the unbiased measure of generalisation performance. 
\end{enumerate}

\subsection{Classifier pre-training} \label{sec:classifier_pretrain}

Before we train our generative model, we need to train a classifier which can easily be adapted to any new classes we encounter. To do this, we first need to train it on a large dataset where there exists many examples per class, so that the features learned are robust and reliable. In our case, this is the training set $\dtrain$.

The classifier we train is a ResNet-50 \citep{he2016deep} initialised from scratch, trained with ADAM \citep{adam} using learning rate $1 \times 10^{-4}$ and moving average coefficients $\beta = (0.9, 0.999)$. The training set is split into a 95\%-5\% split, with the latter forming an internal validation set for early stopping. We train this classifier with a moderate amount of data augmentation, which can comprise randomly-resized crops anywhere between 70\% to 100\% of the original image size, and random rotations anywhere between -10 and 10 degrees. We simply train the classifier using the small internal validation set as an early stopping criteria.

\subsection{Generative adversarial networks} \label{sec:gan_training}

The generative adversarial network \citep{jsgan} we use here is based on the projection discriminator proposed by \cite{cgan_proj}, with a slight modification to make it more similar to the recently proposed StyleGAN class of models \citep{styleganv1}. In StyleGAN, rather than the latent code being fed directly as input to the generator network, the input is instead a learnable constant tensor $\bm{h}_0$ whose progressively growing representation is modulated by the latent code $\bm{z}$ via adaptive normalisation layers. This decision was made to more easily facilitate fine-tuning to new classes, since the adaptive normalisation layers comprise relatively few parameters compared to the main backbone. However, it has also been shown in \citet{styleganv1} that this particular architecture is superior to traditional architectures in terms of image quality and disentanglement.

\begin{figure}[h]
	\centering
	\begin{subfigure}[b]{0.4\textwidth}
		\centering
		\includegraphics[width=\textwidth]{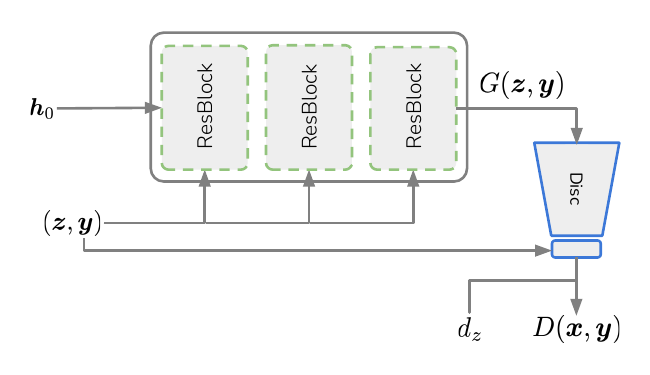}
		\caption{}
		\label{fig:g_arch}
	\end{subfigure} \ \ 
	\begin{subfigure}[b]{0.4\textwidth}
		\centering
		\includegraphics[width=\textwidth]{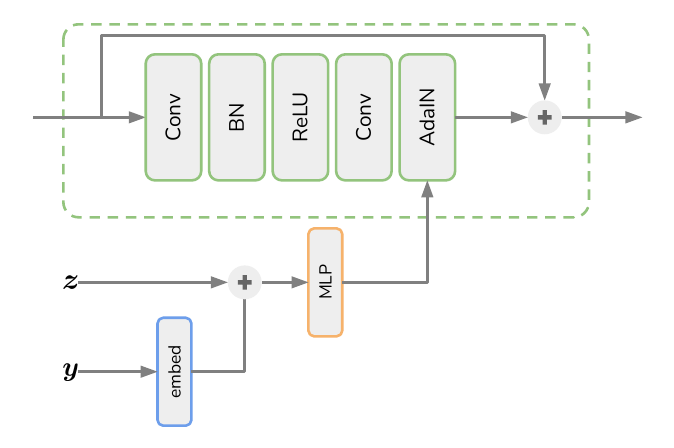} 
		\caption{}
		\label{fig:g_resblock}
	\end{subfigure}
	\caption{\textbf{Left:} overview of both the generator and discriminator. \textbf{Right:} The residual block used for the generator. The label $\bm{y}$'s embedding is extracted and concatenated with the latent code to produce scale and shift coefficients for the second (last) batch normalisation layer. The addition symbol for the $(\bm{z},\bm{y})$ part of the architecture indicates tensor concatenation.}
	\label{fig:architecture}
\end{figure}

We train a conditional GAN which comprises two networks: the generator $G(\bm{z}, \bm{y})$ and discriminator $D(\bm{x}, \bm{y})$, where $\bm{z} \sim p(\bm{z}) = \mathcal{N}(0, \bm{I})$ is the latent code and $\bm{y}$ is the label. In terms of label conditioning, we utilise embedding layers that map indices from the label distribution $p(\bm{y})$ to codes which are concatenated with the latent codes $\bm{z}$, which in turn are used to modulate residual blocks in the generator via adaptive instance normalisation. This is illustrated in Figure \ref{fig:architecture}. The objective function we use is the non-saturating logistic loss originally proposed in \cite{jsgan}:
\begin{align} \label{eq:gan_eqns}
	\min_{D} \mathcal{L}_{D} & = - \mathbb{E}_{\bm{x}, \bm{y} \sim p(\bm{x}, \bm{y})} \log\big[ D(\bm{x}, \bm{y}) \big] - \mathbb{E}_{\bm{z} \sim p(\bm{z}),  \bm{y} \sim p(\bm{y})} \log\big[ 1 - D(G(\bm{z}, \bm{y}), \bm{y}) \big] \\
	\min_{G} \mathcal{L}_{G} & = - \mathbb{E}_{\bm{z} \sim p(\bm{z}), \bm{y} \sim p(\bm{y})} \log\big[ D(G(\bm{z}, \bm{y}), \bm{y}) \big],
\end{align}
where $D = \sigm(d(\bm{x}, \bm{y}))$, with $d()$ denoting the raw logits of the output. In practice, we experienced an issue where the GAN would exhibit serious mode dropping. This appears to be because of our use of a learnable constant as input to the network, though it is currently unclear what architectural choices in StyleGAN mitigate this. To fix this, we utilise the InfoGAN \citep{infogan} loss where the discriminator has to also predict the latent code $\bm{z}$ from the generated image $G(\bm{z}, \bm{y})$, which can be used to maximise mutual information between the generated image and $\bm{z}$. This extra loss is $\| d(G(\bm{z}, \bm{y}))_z - \bm{z} \|^{2}$ and is minimised wrt both networks and is weighted by coefficient $\gamma$, where $d(\cdot)_z$ is a $\bm{z}$ prediction branch that stems off the main backbone of $d$.\footnote{We suspect this loss may have a similar effect to the minibatch standard deviation layer in StyleGAN, which allows the discriminator to distinguish between real/fake based on minibatch statistics.}

\subsection{Fine-tuning} \label{sec:gan_finetuning}

When GAN training has completed, we need to figure out how to best fine-tune it on our support set, which contains very few examples (only $k$) per class. Here, our fine-tuning still involves optimising the adversarial losses in Equation \ref{eq:gan_eqns} as opposed to only fine-tuning the generator with non-adversarial losses (such as in \citet{improve_sample_div_gan}). However, if we do not restrict the number of parameters involved during optimisation this can easily lead to overfiting. Since we are using the projection discriminator of \cite{cgan_proj}, the discriminator's output logits can be written as $d(\bm{x}, \bm{y}) = \bm{y}^{T}\bm{V} \cdot \phi(\bm{h}) + \psi(\phi(\bm{h}))$, where $\bm{h} = f_{D}(\bm{x})$ is the output of the discriminator's backbone and $\bm{V}$ is the embedding matrix that maps a (one-hot encoded) label to its embedding. From this equation, we can devise three ways in which $D$ could be finetuned, and we call this hyperparameter $\dfm$ (`D finetuning mode'): \emph{dfm=embed} corresponds to only updating $\{\bm{V}, d(\cdot)_z\}$; \emph{dfm=linear} corresponds to updating $\{\bm{V}, \phi, \psi, d(\cdot)_z\}$; and \emph{dfm=all} corresponds to updating the backbone $f_D$ as well. We can also define the `G finetuning mode' (\gfm) in a similar manner, with \emph{gfm=embed} denoting that we only update the embedding matrix per residual block, and \emph{gfm=linear} denoting both the embedding matrix and the MLP which produces adaptive instance norm parameters (see Figure \ref{fig:g_resblock}). These choices will be compared against each other in Section \ref{sec:exps_and_results}.

\section{Related work}

A very closely related work is DAGAN \citep{dagan}. The authors proposed training a GAN where the discriminator is trained to distinguish pairs of images, rather than images conditioned on labels.. This means that $D$ is trained to distinguish between a same-class tuples from the real distribution $(\bm{x}_1, \bm{x}_2)$ and a pair from the fake distribution, where one of the elements is generated $(\bm{x}, \tilde{\bm{x}})$, where $\tilde{\bm{x}} = G(r(\bm{x}), \bm{z}))$. In this case, $r(\bm{x})$ is an encoder network and $G$ is the corresponding decoder. The authors motivate this by wanting a discriminator that can naturally generalise to new classes, since it is not explicitly conditioned on a label. For their generator network, $r(\bm{x})$ is meant to infer (without any explicit label supervision) the `content' of the image, while $\bm{z}$ serves as the `style' which is simply noise injected from a prior.  
However, upon attempting to reproduce this model we found a major deficiency, in that the trained decoder $G$ ignores $r(\bm{x})$ completely, i.e $G(\bm{r}, \bm{z}) \approx G(\bm{0}, \bm{z})$. This seems most likely attributable to the fact there is no supervision at all for $r(\bm{x})$, i.e. a reconstruction error or InfoGAN-style losses to ensure that both $\bm{z}$ and $r(\bm{x})$ are used by the network. Such losses do not appear to be present the paper. Furthermore, from the results it is unclear if the method would outperform a discriminator that conditions on the class label for the training set. Despite this, a big difference between their works and ours is that they try to perform zero-shot generation with respect to the \emph{generative model}, since it itself is not fine-tuned on the novel classes. In our own preliminary experiments however a model we trained similar to DAGAN, we found that when we tried to generate images from novel classes with $r(\bm{x})$, there was a strong bias towards it generating images that looked like classes from the training set.


F2GAN \citep{f2gan} proposes an adversarially-augmented autoencoder where $k$-way mixup \citep{mixup, manifold_mixup} is performed between the latent features of $k$ images, coupled with an attention module to patch up discrepencies in the image (hence `fusing and filling'). Unlike DAGAN, a conditional discriminator is used which conditions on labels. In F2GAN, a `mode seeking' loss is proposed to ensure that, for the same pairs of images, different mixing coefficients correspond to sufficiently diverse images. While an ablation study was performed to justify each component/loss that was added, this was not done for any of the datasets for which few-shot classification was performed and therefore it is unclear which components of the model are most influential. Like DAGAN, there is no finetuning on the support set of novel classes, and the model is expected to perform zero-shot generation. We do not pursue zero-shot generation in this paper because we found it is extremely difficult to expect GANs to generalise to unseen classes, a concern that is also shared by \citet{augintae}. It is likely that the multitude of losses proposed in F2GAN are there to inject strong priors into the network such that it can perform well for zero-shot generation, but in this work we take a more direct approach where we fine-tune our GAN on few-shot classes. (Additional discussion regarding zero-shot generation for GANs (and VAEs) can be found in Section \ref{sec:appx_related}.)


One major issue in comparing these papers' results is that there is no standardised way in which the data is split for training and evaluation. DAGAN is the most rigorous in its evaluation, utilising a `source' domain (training set), a `validation' domain (validation set), and a `target' domain (test set). For the validation and testing sets, $k$ examples per class are held out in each, which can be thought of as the support sets for both the validation and testing sets, respectively. These support sets are leveraged by the generative model to create more examples for their respective classes. In DAGAN, the validation set is used to tune the number of synthetic examples that should be generated per class. Conversely, in F2GAN there appears to be no mention of a validation set, and the data is simply split between a training domain and a testing domain. Beause of hyperparameter tuning, it is likely performance estimates are biased. This will be discussed in further detail in Section \ref{sec:exps_and_results}. 

There are many works which involve the fine-tuning of GANs. \citet{robb2020few} proposes a new method to finetune a pre-trained StyleGAN \citep{styleganv1} in the few-shot regime by tuning only the singular values of both networks. \citet{styleganv2_ada} propose a trick to mitigate discriminator overfitting to improve StyleGAN2 training, and present transfer learning results in the context of domain adaptation. Such tricks could be leveraged to potentially improve our results in Section \ref{sec:gan_finetuning}. The aforementioned work of \cite{augintae} also performs few-shot data augmentation but in the context of domain adaptation, where one is interested in mapping between datasets. \cite{icgan} propose `instance conditioned' GANs that condition on pre-trained self-supervised embeddings rather than class labels, and demonstrate impressive zero-shot transfer to other datasets, simply by computing embeddings over these other datasets. While these works undoubtedly present impressive results on larger and more ambitious datasets, none of them have examined a few-shot pipeline where the goal is to generate images to improve a downstream task directly, such as classification performance. These works instead primarily report image similarity metrics computed on their generated images.

A closely related subfield is that of few-shot hallucination \citep{hariharan2017low}, which may also involve using generative models \citep{li2020adversarial} to `hallucinate' (generate) additional examples for novel classes, but in latent space instead of input space. Similar to our procedure, these hallucinated latent samples are combined with real latent samples to make one larger dataset. In principle, our method could be adapted to generate latent codes (with respect to a latent distribution), but this is left to future work. Lastly, while we do not consider meta-learning here, our few-shot pipeline could be framed as an end-to-end approach under that paradigm. There are numerous works combining generation with few-shot classification~\citep{metagan, wang2018low, chen2019image, verma2019meta}.

\section{Experiments and Results} \label{sec:exps_and_results}

We run experiments over five randomised dataset seeds (splits) on the balanced version of the EMNIST dataset \citep{emnist}. For each dataset seed, the training set and validation sets comprise of 38 and 9 randomly sampled classes (without replacement), respectively. Each class comprises 2800 examples, with 80\% being allocated for validation and 20\% for testing. The support set is a subset of the validation set, with only $k$ examples per class. The test set shares the same classes as the validation set but is held out and is only used for unbiased evaluation for classifier fine-tuning in Section \ref{sec:results_cls_finetuning}. Next, we summarise our findings with regards to the stages proposed in Section \ref{sec:proposed}.

\paragraph{GAN pretraining (Section \ref{sec:gan_training}):} We found that our InfoGAN coefficient $\gamma$ does not have to be anywhere close to zero for it to have its intended effect of mitigating mode dropping. We train with ADAM \citep{adam} parameters $\beta = (0.0, 0.9), \text{lr}=2 \times 10^{-4}$, InfoGAN coefficient $\gamma = 100$, with a G:D update ratio of 1:5. For early stopping, we monitor the FID \citep{ttur} between our generated samples and training set. More specifically, we randomly sample $N=5000$ examples from the training set to compute the statistics of the reference distribution, and with the labels extracted from those examples we conditionally generate samples from our GAN which comprises the generative distribution. This metric serves as our early stopping criteria.

\begin{figure}[h!]
	\centering
	\begin{subfigure}[b]{0.397\textwidth}
		\centering
		\includegraphics[width=\textwidth]{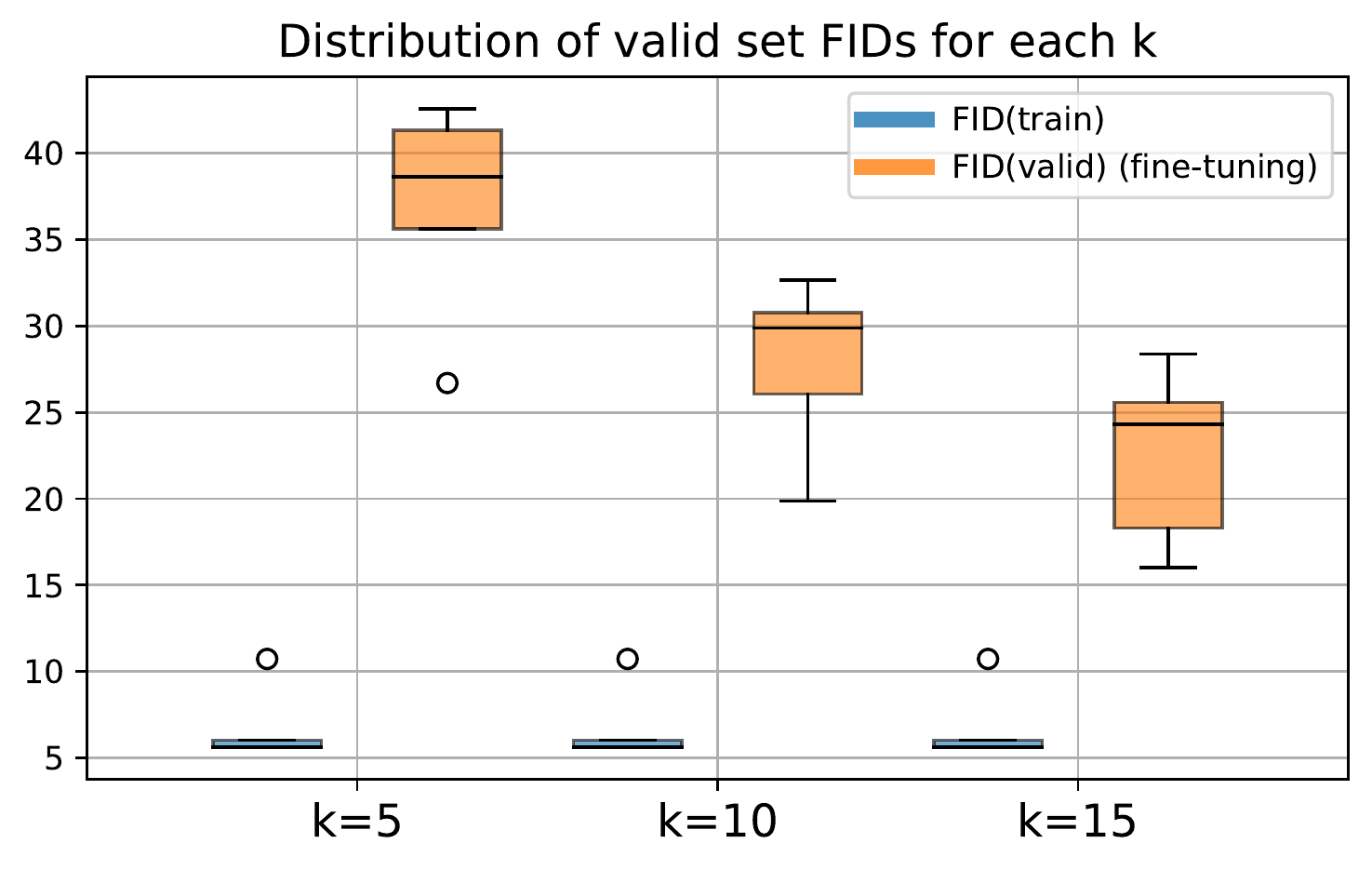}
		\caption{}
		\label{fig:boxplot_ft_a}
	\end{subfigure}
	\begin{subfigure}[b]{0.4\textwidth}
		\centering
		\includegraphics[width=\textwidth]{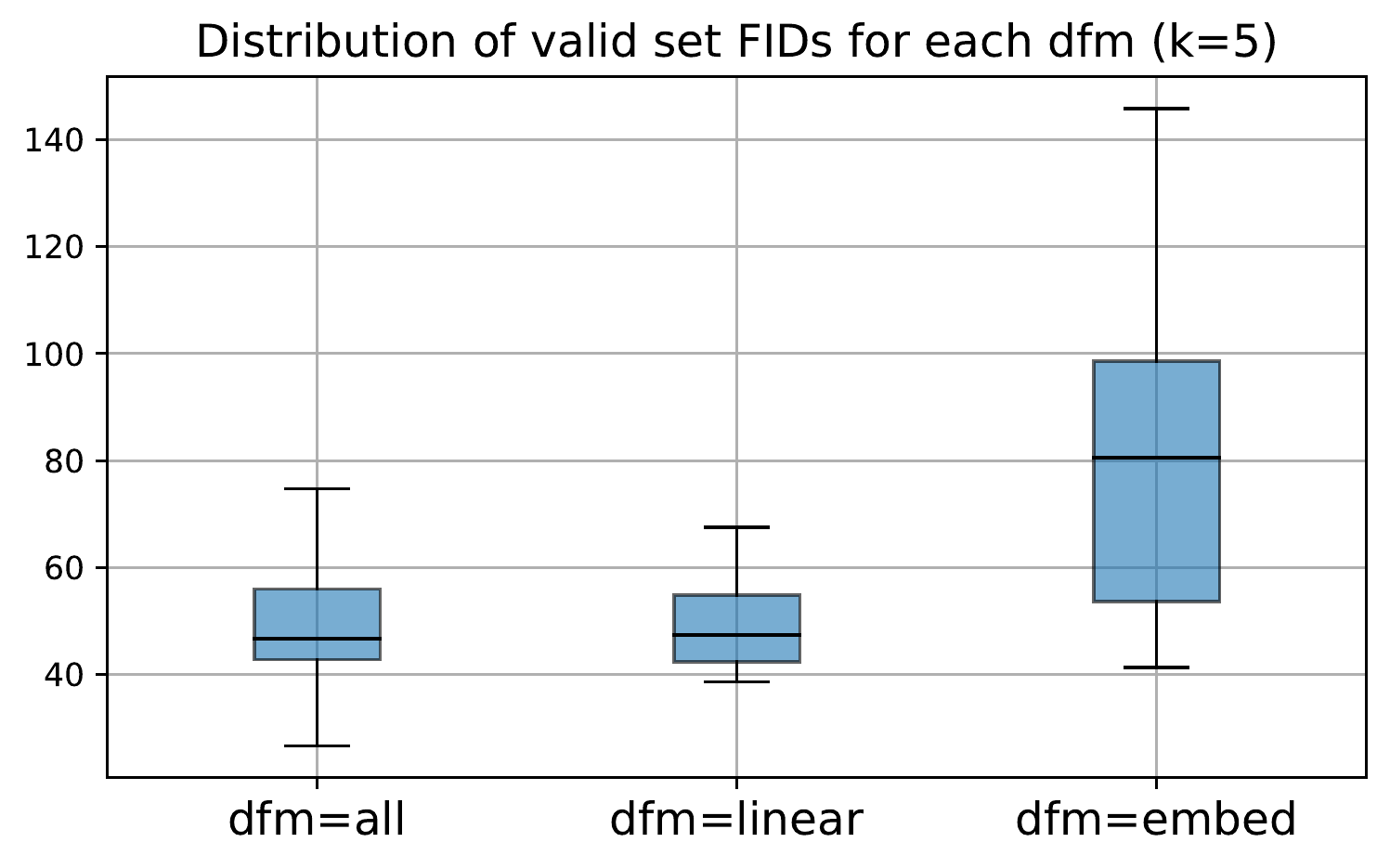}
		\caption{}
		\label{fig:boxplot_ft_b}
	\end{subfigure}
	\caption{\textbf{Left}: For each $k$, we show the distribution of FID valid set scores (over dataset seeds) when the GAN is fine-tuned on the support set $\dsupp$. For each of these $k$, the training set FID is shown as blue as a reference. \textbf{Right}: the distribution of FID valid set scores (over dataset seeds and \gfm modes) for different \dfm modes.}
	\label{fig:boxplot_ft}
\end{figure}

\paragraph{GAN finetuning (Section \ref{sec:gan_finetuning}):} For our fine-tuning strategy, we found that the best strategy is to optimise \emph{only the embedding layers} in $G$ and, quite surprisingly, to optimise for \emph{all} parameters in $D$, using the same adversarial losses and the same InfoGAN coefficient $\gamma$. In order to facilitate new classes, all that is required is for new indices (rows) to be added to each and every embedding layer's matrix in both networks.\footnote{Implementation-wise, this is not neccessary as we simply assume that we know the total number of classes beforehand (train+valid+test) and initialise the embedding matrix to have precisely this many rows.} Here, our early stopping criteria is the FID on the \emph{validation set}; that is, we randomly sample $N$ examples from the validation set as our reference set and use the labels from those examples for the generative distribution. In Figure \ref{fig:boxplot_ft_a} we show the distribution of valid set FIDs as a function of $k$. Perhaps not surprisingly, as $k$ gets larger we are able to generate samples that progressively become more representative of those in the validation set. Here, we also show the FID on the training set (before fine-tuning) as a reference.

For the sake of comparison, we also measure performance across different \emph{dfm}'s. In Figure \ref{fig:boxplot_ft_b} we plot FID on the validation set for each of the three \dfm modes, with the distribution of FIDs being computed over both dataset seeds and \gfm. We can see that \emph{dfm=embed} performs the worst, and surprisingly, not much difference between \emph{dfm=all} and \emph{dfm=linear}. We find this to be an interesting result given it is the opposite of `Freeze-D' \citep{freeze_d}, whose work found the best results in fine-tuning \emph{all} of $G$ and just the later layers of $D$. 

\begin{figure}[h!]
	\centering
	\begin{subfigure}[b]{0.4\textwidth}
		\centering
		\includegraphics[width=\textwidth]{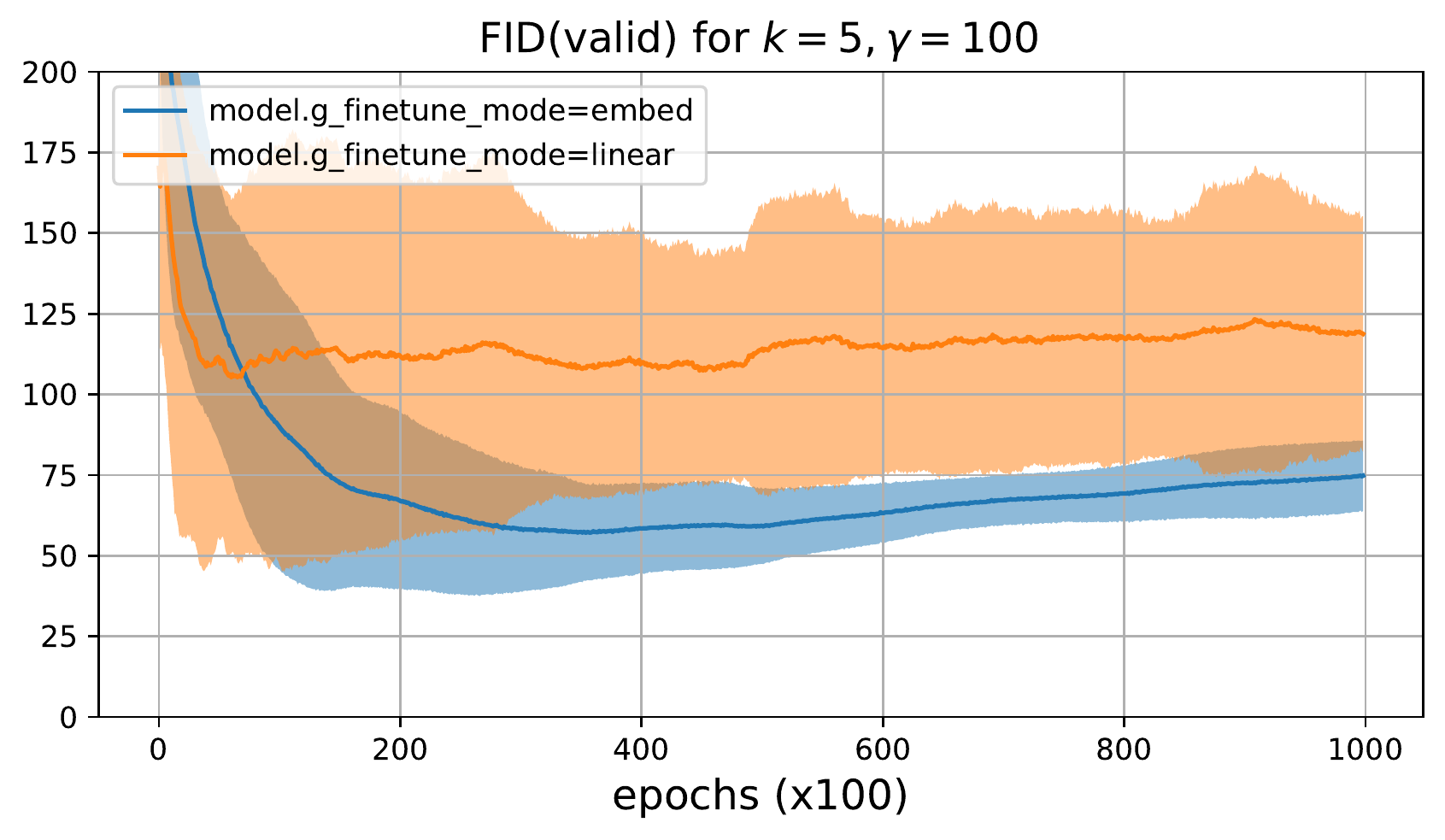}
		\caption{$\gfm \in \{\text{linear}, \text{embed}\}$, valid FID}
		\label{fig:2b_fid_k5_gfm_valid}
	\end{subfigure}
	\begin{subfigure}[b]{0.395\textwidth}
		\centering
		\includegraphics[width=\textwidth]{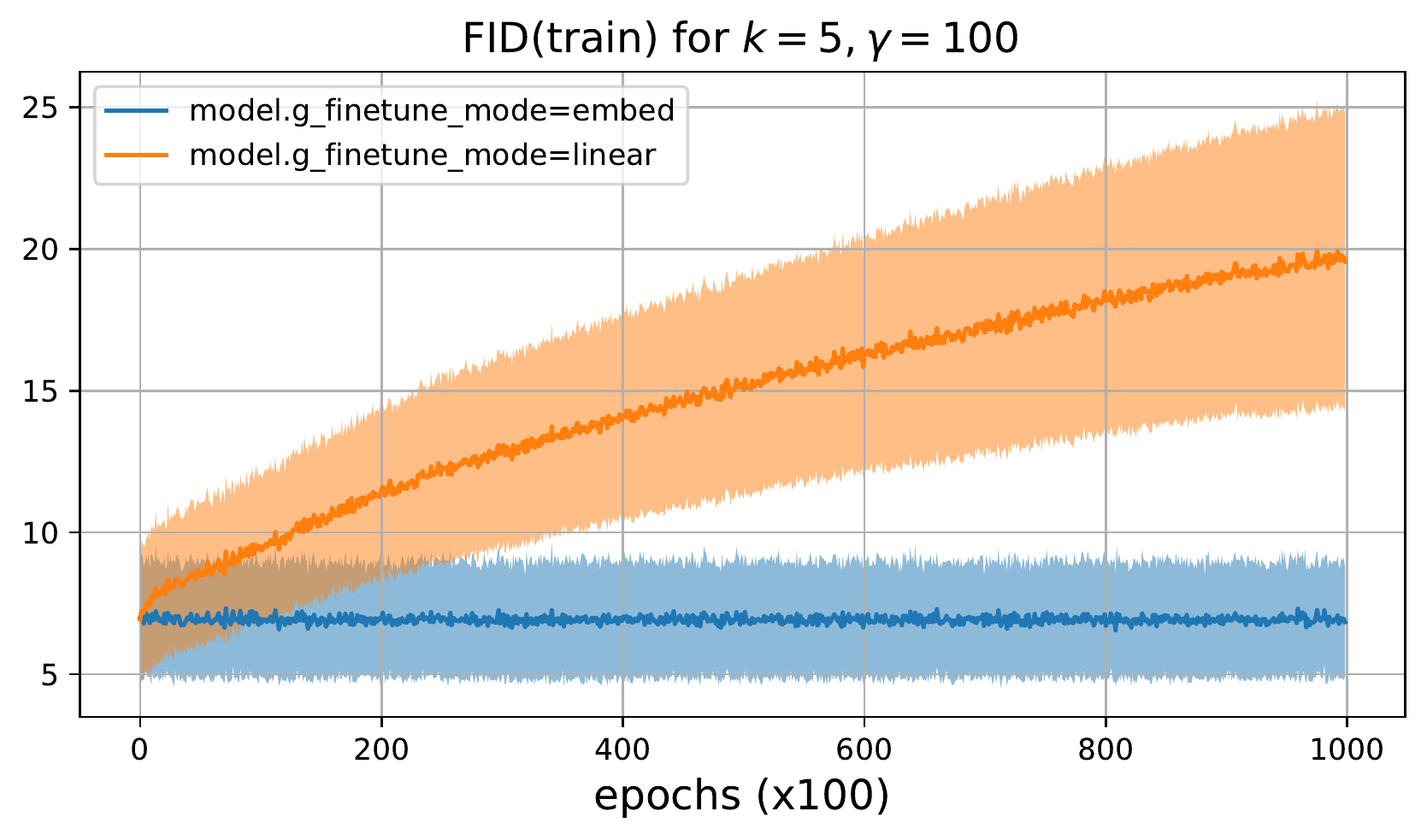}
		\caption{$\gfm \in \{\text{linear}, \text{embed}\}$, train FID}
		\label{fig:2b_fid_k5_gfm_train}
	\end{subfigure}    
	\caption{\textbf{Left}: FID between generated samples on the target classes and the validation set, for GAN fine-tuning (lower FID is better). The orange curve corresponds to our proposed fine-tuning mode for $G$ (embedding layers only, i.e. \emph{gfm=embed}) versus also fine-tuning the MLPs which produce adaptive batch norm coefficients (i.e. \emph{gfm=linear}, see Figure \ref{fig:g_resblock}). Shaded regions (variance) is over all possible \emph{dfm}'s (\emph{all}, \emph{linear}, or \emph{embed}). \textbf{Right}: FID but computed between samples on the source classes and the training set. Here, we can see that fine-tuning only the embedding layers in $G$ does not result in catastrophic forgetting in how to generate samples for source classes.}
	\label{fig:2b_fid_k5}
\end{figure}

In Figure \ref{fig:2b_fid_k5} we present some additional results comparing the fine-tuning mode of $G$, which we denote as \emph{gfm}. We compare fine-tuning just the embeddings in $G$ (\emph{gfm=embed}) versus also fine-tuning the MLPs that take part in adaptive batch normalisation (\emph{gfm=linear}, see Figure \ref{fig:g_resblock}). It can seen that with \emph{gfm=embed}, validation set FIDs are superior on average and that the training set FID does not degenerate over time (Figure \ref{fig:2b_fid_k5_gfm_train}). One may wonder in what situations does it simply suffice to only finetune the embedding layers in $G$ and have it perform well for generation. We recognise that this is highly dependent on both the architecture and datasets used. Unlike the more ambitious task of domain transfer, here our source and target classes can be seen as coming from the same overall distribution $p(\bm{x})$. If source classes in the training set share very similar factors of variation to target classes in the validation or test set, then we would expect the network to generalise to those new classes with minimal modification to the generator network.

\subsection{Classifier finetuning} \label{sec:results_cls_finetuning}

Here, we present the results of our classifier fine-tuning experiments on both the validation and test sets. While we already have a fine-tuned GAN which can generate images from the novel classes, we also need to fine-tune the classifier which was originally trained on the source classes (Section \ref{sec:classifier_pretrain}). To do this, we simply freeze all of its parameters and replacing its output (logits) layer with a new layer which defines a probability distribution over either the validation classes $\yvalid$ or the test classes $\ytest$. Note that unlike with generalised few-shot learning, here we are not interested in maximising performance over both the old and new classes -- we simply wish to maximise performance over the latter. Here, we use the same parameters for ADAM as described in Section \ref{sec:classifier_pretrain} but lower the learning rate to $2 \times 10^{-5}$, and train for 30k epochs. We describe our experiments and their corresponding hyperparameters as follows:

\begin{itemize}[leftmargin=1cm]
	\item \textbf{Baseline}: we fine-tune the pre-trained classifier on $\dsupp$. We run two experiments for this, one with and without traditional data augmentation applied. For the data augmentation experiment, we tune over the minimum random resized crop size \texttt{minScale} $\in \{0.2, 0.4, 0.6, 0.8\}$. 
	\item \textbf{Mixup}: use mixup in input space \citep{mixup}, where the mixing distribution $\text{Beta}(\lambda, \lambda)$ is a hyperparameter, and $\lambda \in \{0.1, 0.2, 0.5, 1.0\}$. This is done `on the fly': for each minibatch seen during fine-tuning, input mixup is applied to both the images and their labels. \texttt{minScale} is fixed to 0.8 here.
	\item \textbf{Fine-tuned GAN}: Using the fine-tuned GAN, we pre-generate $n_s$ new samples per class and combine these with the original supports $\dsupp$ to create $\daug$. Here, $n_s \in \{2, 5, 10, 20\}$ is a hyperparameter that we explore. \texttt{minScale} is also fixed to 0.8 here.
\end{itemize}


\begin{figure}[h]
	\centering
	\begin{subfigure}[b]{0.445\textwidth}
		\centering
		\includegraphics[width=\textwidth]{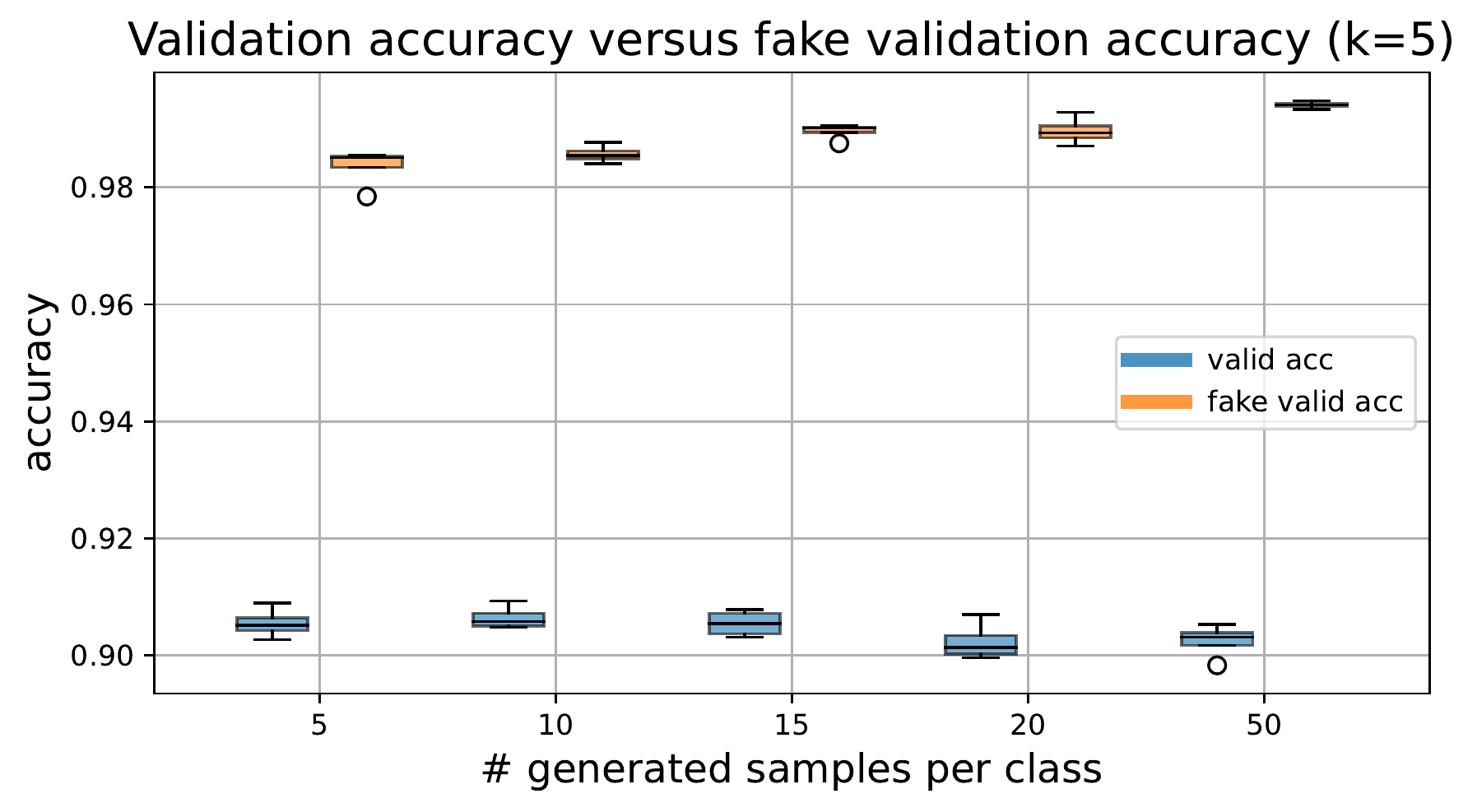}
		\caption{$k=5$}
		\label{fig:boxplot_fakeval_k5}
	\end{subfigure}
	\begin{subfigure}[b]{0.45\textwidth}
		\centering
		\includegraphics[width=\textwidth]{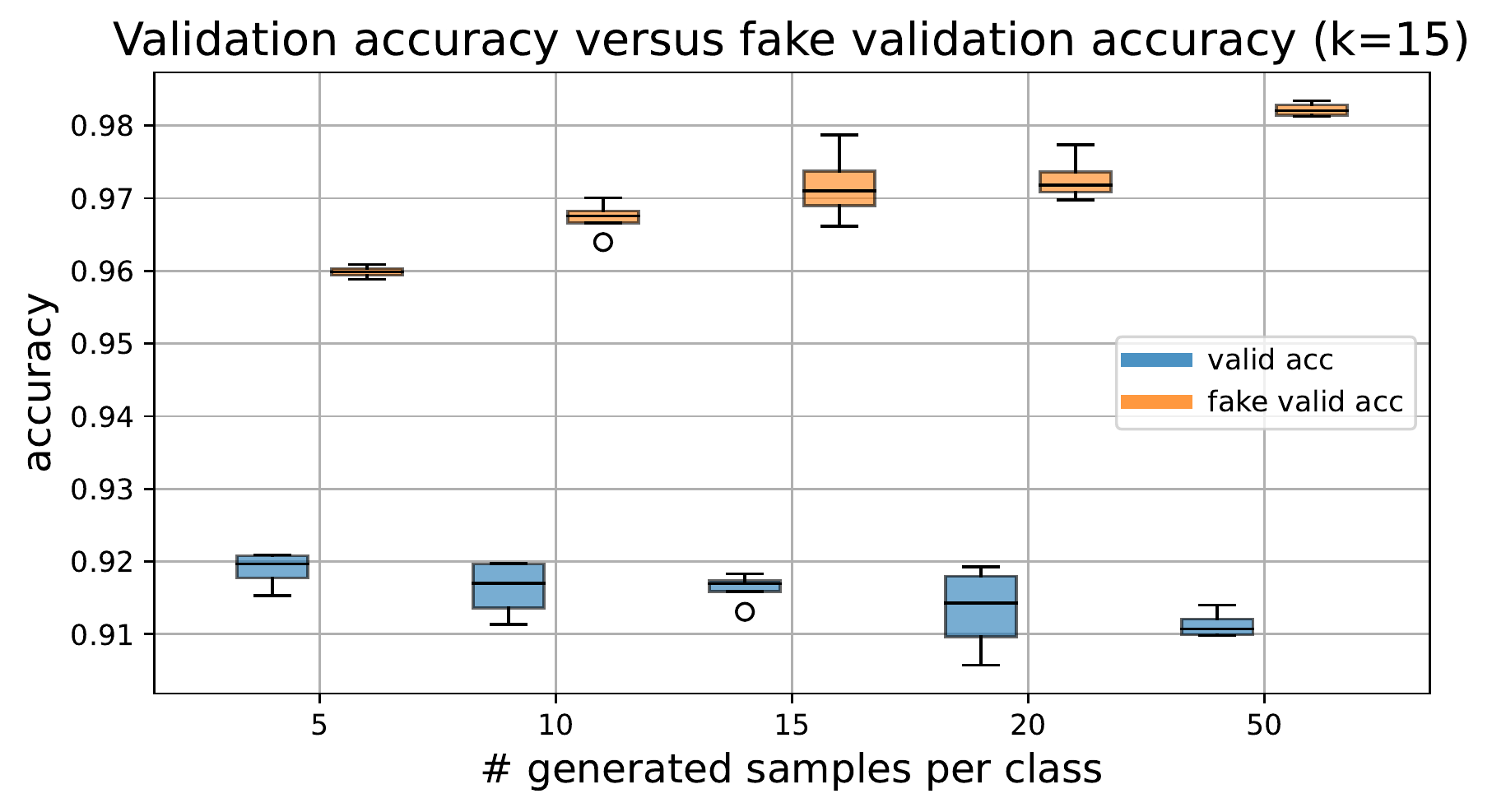}
		\caption{$k=15$}
		\label{fig:boxplot_fakeval_k15}
	\end{subfigure}
	\caption{Boxplot showing validation accuracy (blue) and fake validation accuracy (orange) for different numbers of generated samples per class $n_s$. Fake validation accuracy is a held-out validation set generated by the GAN. It can be seen that as $n_s$ increases, the actual validation accuracy decreases and fake validation accuracy increases. Results are shown over one dataset seed, with each boxplot being a distribution over the different sampling runs.}
	\label{fig:boxplot_fakeval}
\end{figure}

\setlength{\tabcolsep}{12pt}
\begin{table*}[h]
  \caption{Accuracy (\%) of different methods on EMNIST on $\dvalid$. The separating bar distinguishes between results we have lifted from the F2GAN \citep{f2gan} paper and our own results. These two groups of results are not directly comparable due to a variety of confounding factors in the experimental setup. In our results, uncertainty estimates are computed over five different randomised dataset splits (seeds) on the validation set. For each of these seeds, hyperparameter tuning was used to obtain the best result. The best hyperparameters can be found in Section \ref{sec:appx_hps}.}
  \centering
  \begin{adjustbox}{max width=0.8\textwidth}
  \begin{tabular}{P{0.01cm} lccc}
	  \toprule[0.8pt]
	  \multirow{11}{*}{\rotatebox[origin=c]{90}{F2GAN}} & \multirow{2}{*}{Method} &
	  \multicolumn{3}{c}{EMNIST}\cr
	   & & $k=5$  & $k=10$ & $k=15$ \cr
	   \cmidrule(r){2-2} \cmidrule(r){3-3} \cmidrule(r){4-4} \cmidrule(r){5-5} 
	& Baseline & 83.64 & 88.64 & 91.14 \cr
	& \footnotesize{  w/ data aug.}  & 84.62  & 89.63 & 92.07 \cr
	& FIGR \citep{figr} &85.91  & 90.08 & 92.18 \cr
	& GMN \citep{gmn}   & 84.12  & 91.21 & 92.09 \cr
	& DAWSON \citep{liang2020dawson} &  83.63 & 90.72 & 91.83 \cr
	& DAGAN \citep{dagan} &  87.45  & 94.18& 95.58 \cr
	& MatchingGAN \citep{hong2020matchinggan} &  91.75 & 95.91 &96.29 \cr
	& F2GAN \citep{f2gan} & $\textbf{93.18}$& $\textbf{97.01}$ &$\textbf{97.82}$ \cr
	\bottomrule
	\multirow{5}{*}{\rotatebox[origin=c]{90}{ours}} & Baseline & 90.97 $\pm$ 4.62 & 92.68 $\pm$ 4.01 & 93.45 $\pm$ 3.69 \cr
	& \footnotesize{  w/ data aug.} & \textbf{91.27 $\pm$ 4.72} & 92.77 $\pm$ 4.06 & \textbf{93.47 $\pm$ 3.80} \cr
	& Input mixup \citep{mixup} & 91.00 $\pm$ 4.39 & 92.68 $\pm$ 3.99 & 93.32 $\pm$ 3.68 \cr
	& Fine-tuned GAN & 91.11 $\pm$ 4.62 & \textbf{92.85 $\pm$ 4.02} & 93.35 $\pm$ 3.79 \cr
	\midrule 
	& + semi-supervised (Sec. \ref{sec:further_discussion}) & \textbf{92.12 $\pm$ 4.34} & \textbf{93.27 $\pm$ 4.00} & \textbf{93.79 $\pm$ 3.72} \cr
	\bottomrule[0.8pt]
  \end{tabular}
  \end{adjustbox}
  \vspace{0.1mm}
  \label{tb:results}
\end{table*}

\begin{table*}[h]
    \caption{Results of selected experiments on the held-out test set $\dtest$. For each experiment and its dataset seed, the hyperparameters used for the test set evaluation correspond to the optimal hyperparameters found for the same corresponding experiment / seed on the validation set (e.g. \# of epochs, $\alpha$ for input mixup, $n_s$ for GAN).}
    \centering
    \begin{adjustbox}{max width=0.8\textwidth}
    \begin{tabular}{P{0.01cm} lccc}
      \toprule[0.8pt]
       & \multirow{2}{*}{Method} & \multicolumn{3}{c}{EMNIST} \cr 
       & & $k=5$  & $k=10$ & $k=15$ \cr
      \cmidrule(r){2-2} \cmidrule(r){3-3} \cmidrule(r){4-4} \cmidrule(r){5-5} 
      \multirow{3}{*}{\rotatebox[origin=c]{90}{ours}} & Baseline w/ data aug & 90.65 $\pm$ 4.52 & 92.22 $\pm$ 4.19 & 93.02 $\pm$ 3.94 \cr
      & Input mixup \citep{mixup} & 90.69 $\pm$ 4.34 & 92.27 $\pm$ 4.05 & 93.04 $\pm$ 3.65 \cr
      & Fine-tuned GAN & 90.81 $\pm$ 4.53 & 92.50 $\pm$ 4.11 & 92.90 $\pm$ 4.03 \cr
      \midrule 
      & + semi-supervised (Sec. \ref{sec:further_discussion}) & \textbf{91.64 $\pm$ 4.35} & \textbf{92.95 $\pm$ 4.14} & \textbf{93.42 $\pm$ 3.84} \cr
      \bottomrule[0.8pt]
    \end{tabular}
    \end{adjustbox}
    \vspace{0.1mm}
    \label{tb:results_test}
  \end{table*}  

We present our results in in Table \ref{tb:results}. We can see that all results are characterised by non-negligible variances, especially when $k$ is small. This can be problematic because it can diminish the statistical significance of any claims in performance gains. If we just consider the mean accuracy however, the data augmentation baseline performs best for $k=5$ and $k=15$ and our method performs best for $k=10$. We found that validation accuracy is highly dependent on the number of generated samples per class $n_s$, and often times accuracy degrades when it is too high. We conjecture that this is because there is a fundamental mismatch between the distribution of GAN-generated images and those from the validation set (whose accuracy we wish to maximise), and this is leading to overfitting on the former. To validate this, we generate a held-out validation set from the same GAN that we generated the images from, which we call our `fake' validation set. We can see in Figure \ref{fig:boxplot_fakeval} that as $n_s$ increases, so does accuracy on our fake validation set, and this is negatively correlated with accuracy on the actual validation set. This clearly indicates we are indeed overfitting to the distribution induced by our GAN. 

The results reported by F2GAN \citep{f2gan} in the upper panel of Table \ref{tb:results} are indeed notable compared to ours for $k=10$ and $k=15$, but unfortunately their description of how the dataset was split is unclear and there are concerns with how principled this evaluation is. In their work, they describe EMNIST as comprising of 47 classes but with 28 being selected as source (training) classes and 10 as targets (testing), leaving 9 classes unaccounted for. In their preceding work MatchingGAN \citep{hong2020matchinggan}, the dataset is split into 28-10-10 (i.e. train-valid-test) for a total of 48 classes, with 10 classes being used as a validation set for GAN training. The test set (10 classes) is used for classifier evaluation, with a small support set of $k$ examples per class used by the GAN to generate additional examples. However, excessive tuning of hyperparameters of the GAN (that is, the hyperparameters that directly control generation, e.g. the generation seed or number of generated samples per class $n_s$) and hyperparameters of the classifier can lead to biased estimates of performance on the test set, which is why we have also presented results on our test set in Table \ref{tb:results_test}. In this table, we find that for all values of $k$, the data augmentation baseline performs the best in terms of mean accuracy, though like we have mentioned earlier, these results have relatively inflated variances which diminish their significance.

Another confounder is that MatchingGAN report artificially reducing the size of all classes in EMNIST to be 100 examples per class to mimic DAGAN's setup, but it is not clear whether F2GAN has also done this. For reference, the only two splits of EMNIST with 47 classes are \texttt{ByMerge} and \texttt{Balanced}, with \texttt{Balanced} containing 2,800 examples per class and \texttt{ByMerge} containing anywhere between 2,961-44,704 examples per class, so this is an enormous reduction in dataset size. Such a reduction would be detrimental to both classifier pre-training and GAN training. While DAGAN's evaluation protocol differs slightly to our work and the other works, they report 76\% test set accuracy on EMNIST with $k=15$ with this artificial reduction in examples per class. Lastly, for both MatchingGAN and F2GAN, the number of samples generated per class with the GAN was set to $n_s=512$, which is extraordinarily large, considering that we obtain worse results on our validation set with anything more than $n_s=5$, on average. Our results appears to corroborate that of DAGAN, whose authors report only tuning $n_s \in \{1, \dots, 10\}$. Because there are many subtle differences in the empirical evaluation between our work and the aforementioned ones, we simply defer the reader to Figures \ref{fig:appendix_data_splits}, \ref{fig:appendix_matchinggan_setup}, \ref{fig:appendix_dagan_setup} for more information rather than enumerate all of these here.


Lastly, we also present results on Omniglot \citep{omniglot}, where each dataset seed comprises 1411 source classes and 212 target classes. Note however that Omniglot does not fit nicely with our particular training paradigm, which assumes that the source classes are abundant in their number of examples. In Omniglot, while there are many classes (1623), there are only 20 examples per class, which means that we are operating in a low data regime even when we are pre-training our GAN on the source classes. While we achieve the highest validation accuracy here, the unbiased (test set) evaluation favours the data augmentation baseline.

\begin{table*}[h]
    \caption{Results for Omniglot, for $k=5$. Results are averages over five different dataset seeds.}
    \centering
    \begin{adjustbox}{max width=0.6\textwidth}
    \begin{tabular}{P{0.01cm} lcc}
      \toprule[0.8pt]
      & Method & $\dvalid$ & $\dtest$ \cr 
      \midrule
      \multirow{4}{*}{\rotatebox[origin=c]{90}{ours}} & Baseline w/ data aug & 97.35 $\pm$ 0.33 & \textbf{95.56 $\pm$ 0.47} \cr
      & Input mixup & 96.85 $\pm$ 0.34 & 94.95 $\pm$ 0.86 \cr
      & Fine-tuned GAN & 97.33 $\pm$ 0.32 & 95.49 $\pm$ 0.58 \cr
      & + semi-supervised (Sec. \ref{sec:further_discussion}) & \textbf{97.40 $\pm$ 0.29} & 95.28 $\pm$ 0.65 \cr
      \bottomrule[0.8pt]
    \end{tabular}
    \end{adjustbox}
    \vspace{0.1mm}
    \label{tb:results_og}
\end{table*}


\section{Further discussion} \label{sec:further_discussion}

\begin{figure}[h]
	\centering
	\begin{subfigure}[b]{0.45\textwidth}
		\centering
        \includegraphics[width=\textwidth]{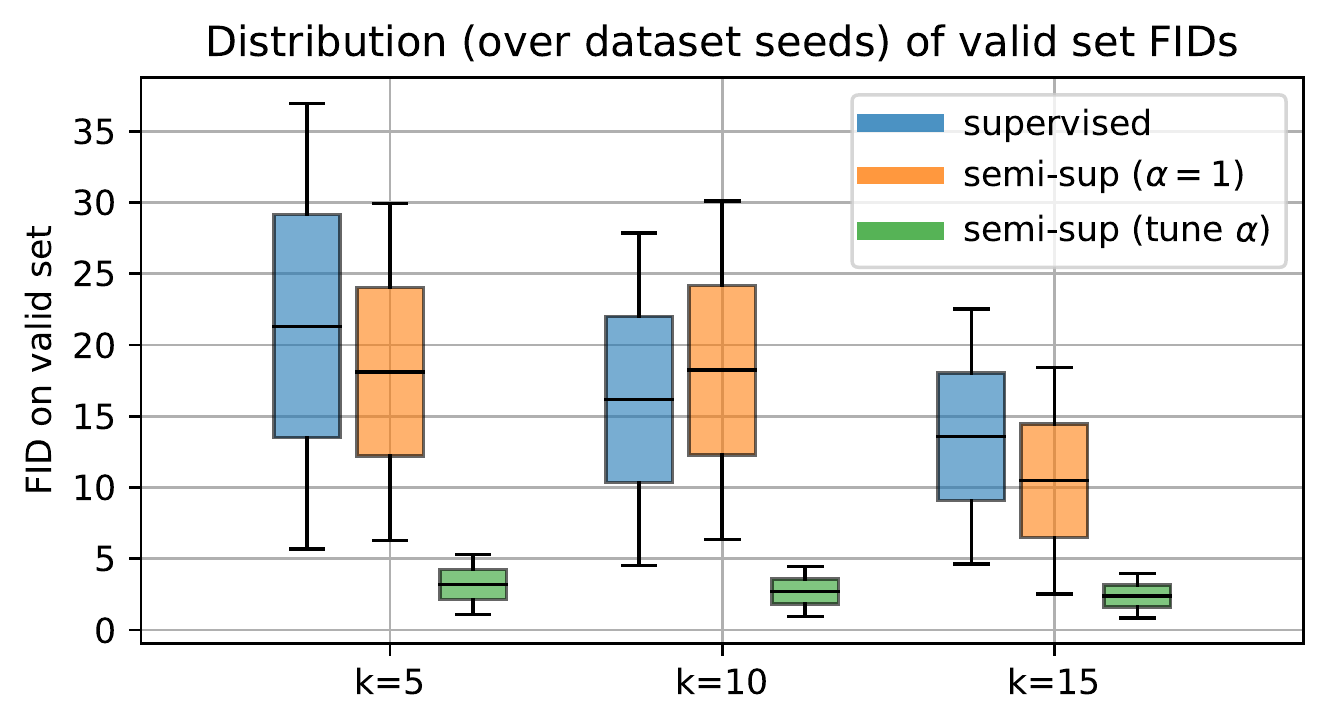}
        \caption{}
        \label{fig:semisup_barplots}
	\end{subfigure}
	\begin{subfigure}[b]{0.45\textwidth}
		\centering
        \includegraphics[width=\textwidth]{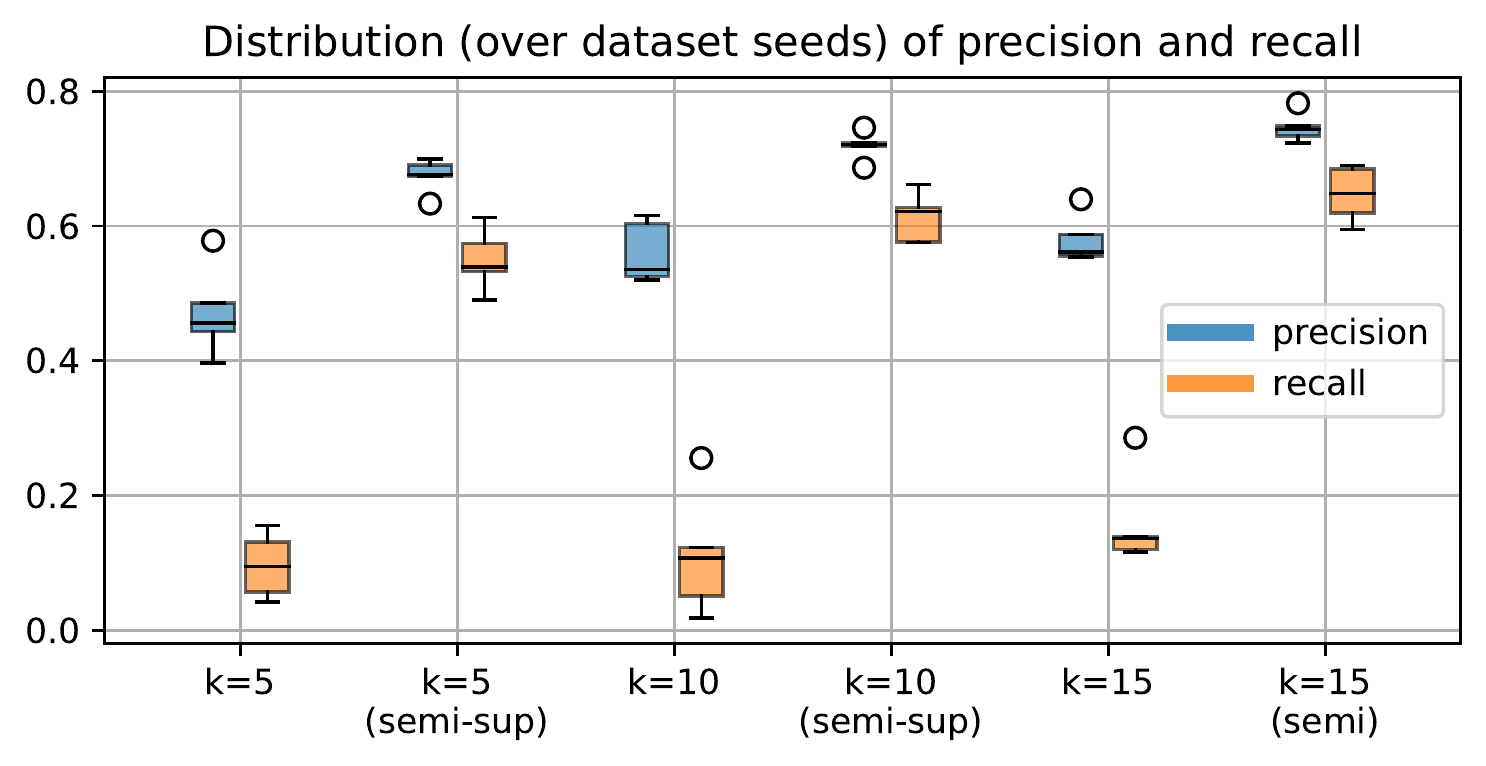}
        \caption{}
        \label{fig:boxplot_pr_emnist}
	\end{subfigure}
	\caption{\textbf{Left:} Comparison of FID scores on $\dvalid$ between supervised (Equation \ref{eq:gan_eqns}) and semi-supervised variant (Equation \ref{eq:gan_eqns_semi}). `Tune $\alpha$' refers to also tuning the $\alpha$ coefficient. \textbf{Right:} Boxplots showing precision-recall metric (proposed in \cite{kynkaanniemi2019improved}) over all $k$'s as well as their semi-supervised variants for EMNIST. We can see that the semi-supervised formulations greatly improve both metrics for each $k$.}
	\label{fig:fd_misc_plots}
\end{figure}

One major limitation of using conditional GANs for data augmentation is that there is no way to leverage unlabelled data, which tends to be more abundant than labelled data. Another issue with our setup is that downstream performance is likely to only give practical gains for a very specific range of values for $k$. For example, when $k$ is too small, there is a strong incentive for one to leverage GAN-based data augmentation since there are very few examples per class, but as we have demonstrated, fine-tuning a GAN is difficult precisely because of how few examples there are. Conversely, as $k$ becomes larger, even if we could finetune a better GAN we would also expect there to be diminishing returns when it comes to improving classification performance over the baseline, since there are an abundant number of labelled examples per class. Therefore, having the ability to leverage unlabelled data seems like a more pragmatic endeavour since we could still experiment with small $k$ but likely do a much better job at fine-tuning a better quality generative model over the new classes. Fortunately, it turns out that one can take the projection discriminator equation of \cite{cgan_proj} that we use and easily turn it into a semi-supervised variant. Recall from Equation \ref{eq:cgan} that the output logits of the discriminator can be written as follows:
\begin{align} \label{eq:cgan}
d(\bm{x}, \bm{y}) = \bm{y}^{T}\bm{V} \cdot \phi(\bm{h}) + \psi(\phi(\bm{h})),
\end{align}
with $\bm{h} = f_D(\bm{x})$, where $f_D$ is the backbone (feature extractor) of the discriminator. From \cite{cgan_proj}, this equation is the result of trying to model the likelihood ratio between the real and generative distributions, which in turn can be decomposed into two ratios that correspond to the terms in Equation \ref{eq:cgan}:
\begin{align} \label{eq:cgan_semi}
d(\bm{x}, \bm{y}) \approx \log \frac{q(\bm{x},\bm{y})}{p(\bm{x},\bm{y})} = \log \frac{q(\bm{y}|\bm{x})q(\bm{x})}{p(\bm{y}|\bm{x})p(\bm{x})} = \underbrace{\log \frac{q(\bm{y}|\bm{x})}{p(\bm{y}|\bm{x})}}_{\approx \bm{y}^{T}\bm{V} \cdot \phi(\bm{h})} + \underbrace{\log \frac{q(\bm{x})}{p(\bm{x})}}_{\approx \psi(\phi(\bm{h}))},
\end{align}
From this, one can easily see that the last term in Equation \ref{eq:cgan_semi} models log ratio $q(\bm{x})/p(\bm{x})$ which is not dependent on any label. From this, we could play an additional adversarial game between $G$ and $D$ for just this term. If we overload notation so that $D(\bm{x}) = \sigm(d(\bm{x})) = \sigm(\psi(\phi(\bm{h})))$ we can derive a semi-supervised set of losses:
\begin{align} \label{eq:gan_eqns_semi}
	\min_{D} \mathcal{L}_{D} & = \mathcal{L}_{D}^{\text{(sup)}} - \alpha \Big[ \mathbb{E}_{\bm{x} \sim p_u(\bm{x})} \log\big[ D(\bm{x}) \big] - \mathbb{E}_{\bm{z} \sim p(\bm{z}),  \bm{y} \sim p(\bm{y})} \log\big[ 1 - D(G(\bm{z}, \bm{y})) \big] \Big] \\
	\min_{G} \mathcal{L}_{G} & =  \mathcal{L}_{G}^{\text{(sup)}} - \alpha \Big[ \mathbb{E}_{\bm{z} \sim p(\bm{z}), \bm{y} \sim p(\bm{y})} \log\big[ D(G(\bm{z}, \bm{y})) \big] \Big],
\end{align}
where $\mathcal{L}_{D}^{\text{(sup)}}$ and $\mathcal{L}_{G}^{\text{(sup)}}$ are their respective terms in Equation \ref{eq:gan_eqns}, $p_u$ denotes some unlabelled distribution (i.e. the validation set but with the labels ignored), and $\alpha$ controls the strength of the unsupervised part of the objective. Note that here the generator is not modified to perform unconditional generation -- it simply has to fool both the conditional branch $d(\bm{x}, \bm{y})$ and unconditional branch $d(\bm{x})$. $\alpha$ should be carefully tuned here, as if it is too large then the generator may fail to generate class-consistent samples. We are not aware of any work which adapts the projection discriminator in this way, though \cite{sricharan2017semi} proposed a semi-supervised CGAN variant which is conceptually similar to Equation \ref{eq:cgan}. To train with this objective, we can simply consider fine-tuning using the entire validation set $\dvalid$ (without labels) for the unsupervised terms in Equation \ref{eq:gan_eqns_semi}, in addition to the labelled support set $\dsupp$ for the supervised terms. In Figure \ref{fig:semisup_barplots} we show the distribution of FID scores obtained when we employ our semi-supervised variant, and also when we are able to tune $\alpha$. We can see that the latter provides a very significant improvement in FID scores, and this improvement in sample quality is also reflected in Table \ref{tb:results}, which achieves the best mean performance across the board. 

Earlier, we explained that mismatches between the generative and real distribution can manifest as degraded accuracy as the number of generated samples increases, and we corroborated this with in Figure \ref{fig:boxplot_fakeval}. To gain further insight into what is going on, we also examine precision and recall metrics \citep{precision_recall,kynkaanniemi2019improved}. Intuitively, precision and recall can be thought of as sample quality and sample coverage, respectively. In Figure \ref{fig:boxplot_pr_emnist} we can see that for each set of experiments the average recall lags behind precision; since recall intuitively corresponds to mode coverage this suggests that few-shot generalisation performance is mostly bottlenecked by the lack of sample diversity. (A similar observation can also be made for Omniglot in Figure \ref{fig:boxplot_pr_og}.)


In general, we consider generative modelling in the few-shot regime to be a difficult task. All generative models present a trade-off \citep{generative_trilemma} between three criteria: sample quality, fast sampling, and mode coverage/diversity, with GANs tending to be problematic in the latter \citep{li2018implicit} since they do not perform maximum likelihood estimation (which favours mode coverage). Variational autoencoders (VAEs) on the other hand suffer from reduced sample quality but at the benefit of mode coverage. Score-based \citep{song2020score} and denoising diffusion probabilistic models (DDPMs) \citep{sohl2015deep, ho2020denoising} appear to perform well in terms of both mode coverage and sample quality, and may be a more appealing alternative since its main bottleneck -- sample generation -- is only of concern if one wants to be able to generate samples in real-time, which is not a concern for us here. Furthermore, as shown in \citep{song2020score} one can also estimate the log density $\log p(\bm{x})$ exactly, which avoids the need to use approximate metrics such as FID. Of course, there is still the issue of training or fine-tuning a generative model on limited amounts of labelled data, but we believe a reasonable approach for the time being is through a semi-supervised approach, since unlabelled data is usually abundant. Lastly, we note that it may be an interesting avenue to instead perform model selection using a weighted combination of precision and recall, since this allows us to weight our preferences for sample coverage and quality with regards to model selection.



\section{Conclusion}

In this paper, we explored the use of conditional generative adversarial networks to perform few-shot data augmentation in order to improve classification performance on various datasets. In general, we found that it is difficult to improve few-shot classification accuracy with a large number of GAN-generated examples due to the inherent mismatch between the ground truth vs generative distribution for the target classes, which leads to overfitting. Part of this is due to GAN intricacies (i.e. the inductive bias to sacrifice mode coverage for sample quality) but this mismatch is heavily exacerbated when there is only $k$ labelled examples per target class. Furthermore, we found that classification performance is highly dependent on how classes are partitioned and this can induce significant variance. To address the difficulty of fine-tuning a GAN to perform $k$-shot generation, we proposed a pragmatic semi-supervised setup that allows for unlabelled examples to be used for GAN fine-tuning and demonstrated gains in fine-tuning accuracy, FID, as well as precision and recall.  The variance exhibited between different dataset seeds is of concern, because it reduces the practical significance of some of the findings. For future work, we suggest building on top of semi-supervised training with more stable and recent generative model classes, such as score-based or diffusion models.

\bibliography{main}
\bibliographystyle{collas2022_conference}

\newpage

\makeatletter
\renewcommand \thesection{S\@arabic\c@section}
\renewcommand\thetable{S\@arabic\c@table}
\renewcommand \thefigure{S\@arabic\c@figure}
\makeatother

\section{Appendix}

\subsection{Additional related work discussion} \label{sec:appx_related}

One crucial difference between what we propose and both of these works is that ours requires the generative model to be fine-tuned on novel classes. \citet{augintae} argue that generative models such as GANs and VAEs have extreme difficulty generalising to novel classes in a `zero-shot' manner (i.e. without fine-tuning) due to the fact that the training of either models enforces that all plausible regions in the space of the prior distribution decode into plausible samples from the training distribution. They argue that unlike their probabilistic variants, \emph{deterministic} autoencoders are fit for few-shot generalisation because of this lack of rigidity, and propose a novel mixup scheme to ensure interpolations between images in novel classes do not produce images biased towards the training set. We argue that in the case of VAEs however, it depends on the trade-off between the two competing losses that comprise the evidence lower bound (ELBO), which is the likelihood + KL (between posterior and prior). The likelihood (reconstruction error) encourages a bijective mapping between $\mathcal{X}$ and $\mathcal{Z}$, which allows it to generalise to novel inputs. The KL loss acts as a regularisation to make the posterior $q(\bm{z}|\bm{x})$ indistinguishable from the prior $p(\bm{z})$, and in the extreme case would bottleneck the capacity of the network and map many distinguishable inputs to the same mode, which is the opposite to the intended goal of the likelihood term \citep{structured_disentanglement}. An extensive literature surrounds the trade-off between these two terms \citep{bvae, structured_disentanglement, mathieu2019disentangling}.  A similar analogy could be made for autoencoders with adversarial losses such as F2GAN, where we can imagine the KL loss being replaced with an `adversarial' divergence between images created by the decoder and images from the training distribution \citep{berthelot2018understanding, sainburg2018generative, amr}. Too much weighting on this term would produce a strong bias in the decoder to ensure \emph{all} images are indistinguishable from those from the training distribution, potentially to the detriment of it being able to generate new samples from inputs coming from unseen classes. This seems to explain why F2GAN contains a multitude of losses to try and encourage sample diversity, though as we have stated, in our work we would prefer to simply fine-tune a GAN to novel classes directly.

\subsection{Additional figures}

\begin{figure}[h]
	\centering
	\begin{subfigure}[b]{0.4\textwidth}
		\centering
		\includegraphics[width=\textwidth]{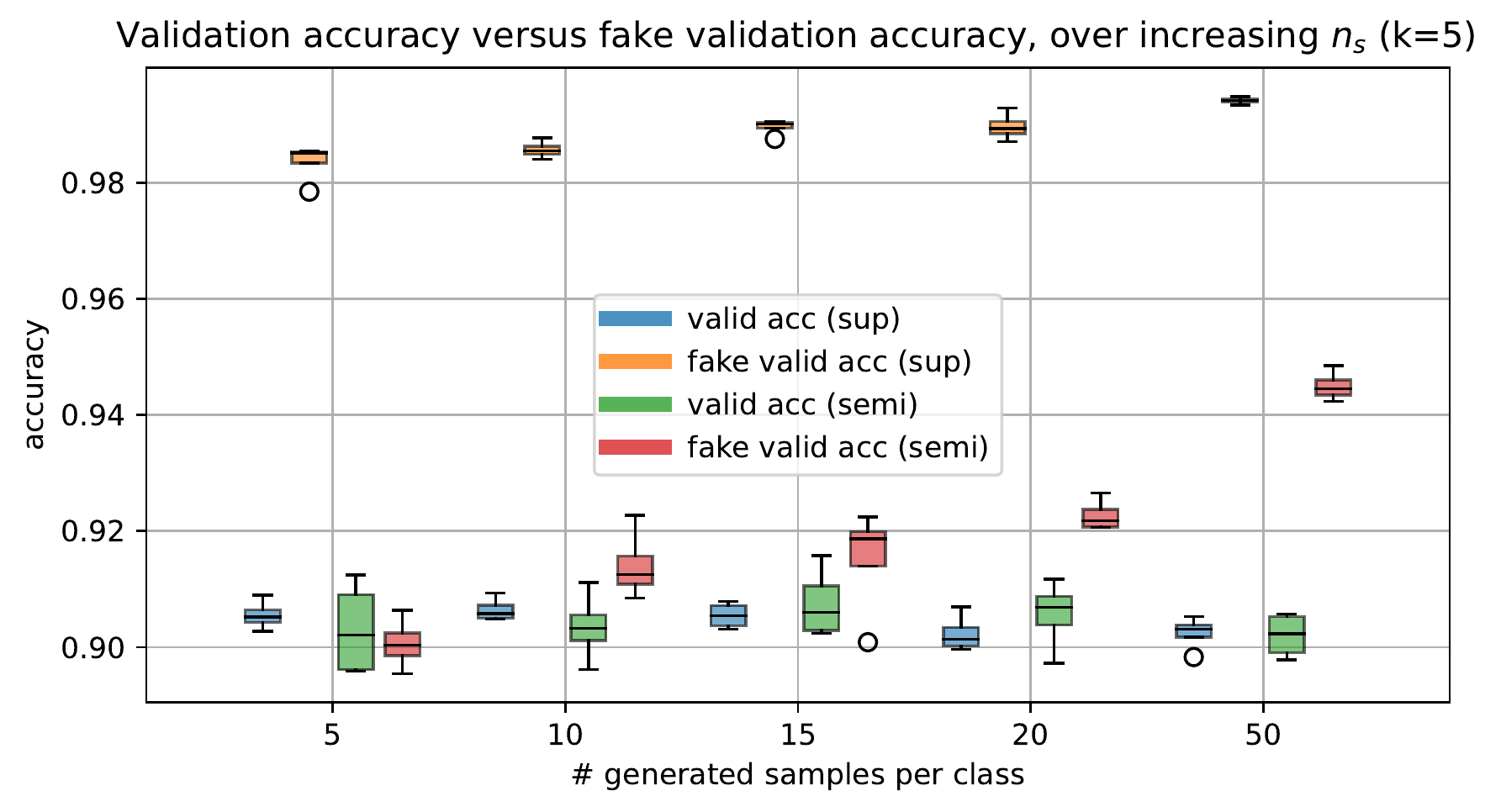}
		\caption{$k=5$}
		\label{fig:boxplot_fakeval_semi_k5}
	\end{subfigure}
	\begin{subfigure}[b]{0.4\textwidth}
		\centering
		\includegraphics[width=\textwidth]{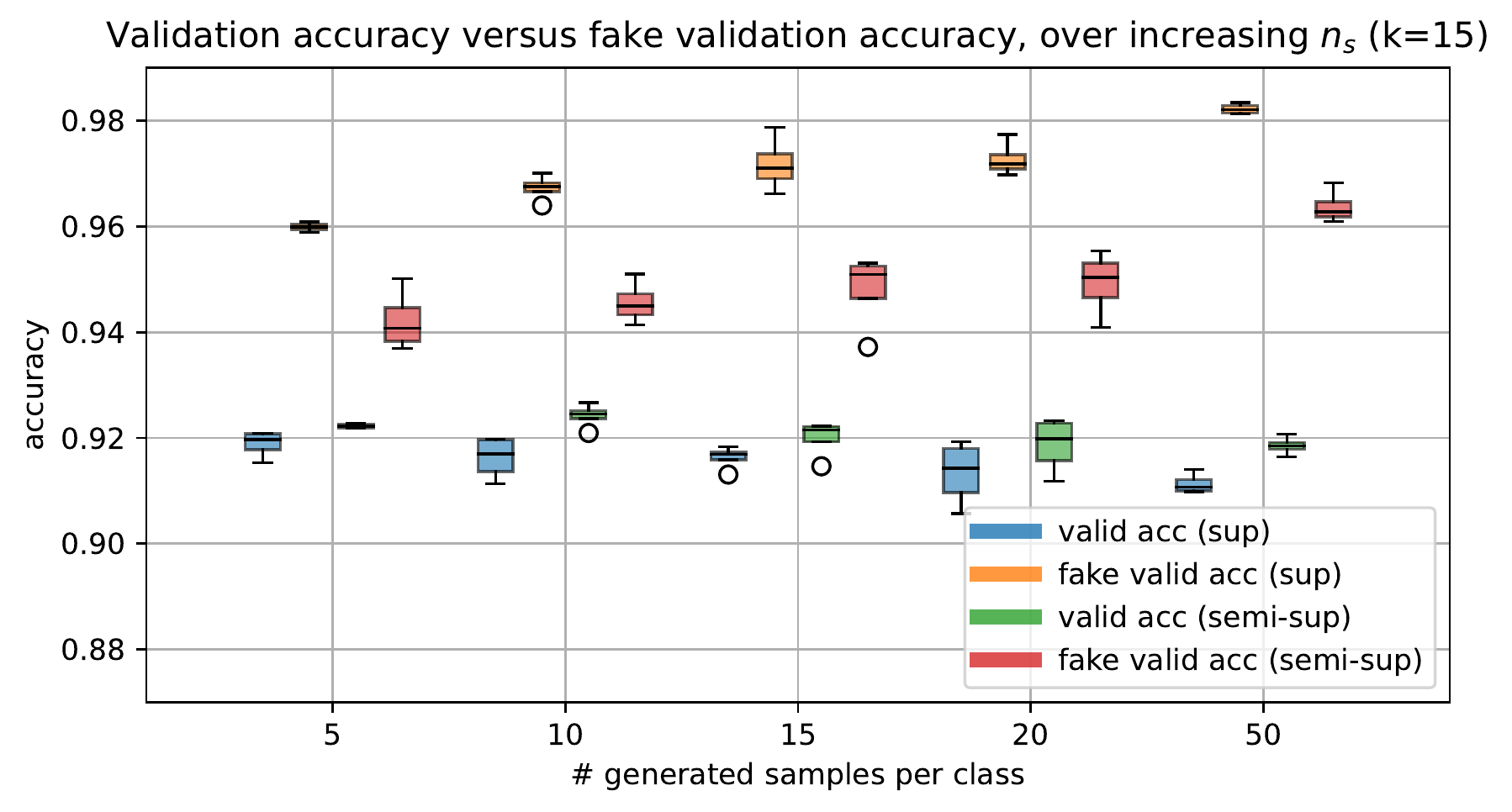}
		\caption{$k=15$}
		\label{fig:boxplot_fakeval_semi_k15}
	\end{subfigure}
	\caption{Boxplot showing validation accuracy (blue = supervised, green = semi-supervised) and fake validation accuracy (orange = supervised, red = semi-supervised) for different numbers of generated samples per class $n_s$. The semi-supervised variant is that described in Equation \ref{eq:gan_eqns_semi}. Fake validation accuracy is a held-out validation set generated by the GAN. It can be seen that as $n_s$ increases, the actual validation accuracy (blue) decreases and fake validation accuracy increases (orange), but this effect is mitigated for the corresponding semi-supervised distributions (shown in green for validation accuracy and red for fake validation accuracy). Results are shown over one dataset seed, with each boxplot being a distribution over the different sampling runs.}
	\label{fig:boxplot_fakeval_semi}
\end{figure}

\begin{figure}[h]
	\centering
    \includegraphics[width=0.6\textwidth]{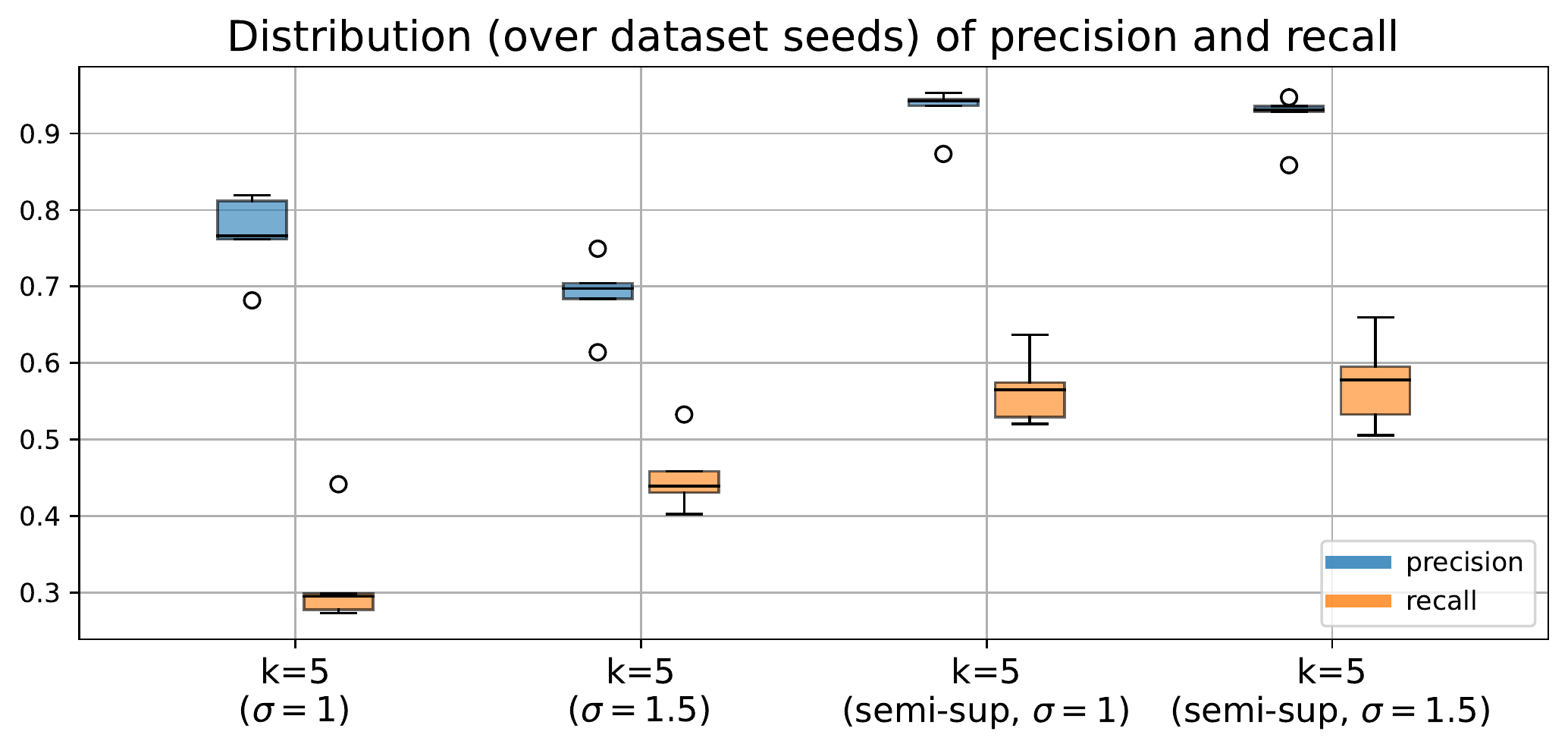}
    \caption{Boxplots showing precision-recall metric (proposed in \cite{kynkaanniemi2019improved}) over all $k$'s as well as their semi-supervised variants for Omniglot. Here, we also illustrate the effect of $\sigma$, which is the standard deviation of the prior distribution at inference time, i.e. $\bm{z} \sim \mathcal{N}(0, \sigma^2 \mathbf{I})$. As demonstrated in BigGAN \citep{biggan}, truncating the dynamic range of $\bm{z}$ leads to increased sample quality at the cost of sample diversity, whereas here we do the opposite by inflating the variance of the prior distribution. We can see that for $\sigma = 1.5$ we obtain better recall (mode coverage) at the expense of precision (sample quality).}
    \label{fig:boxplot_pr_og}
\end{figure}

\begin{figure}[h]
	\centering
    \includegraphics[width=0.4\textwidth]{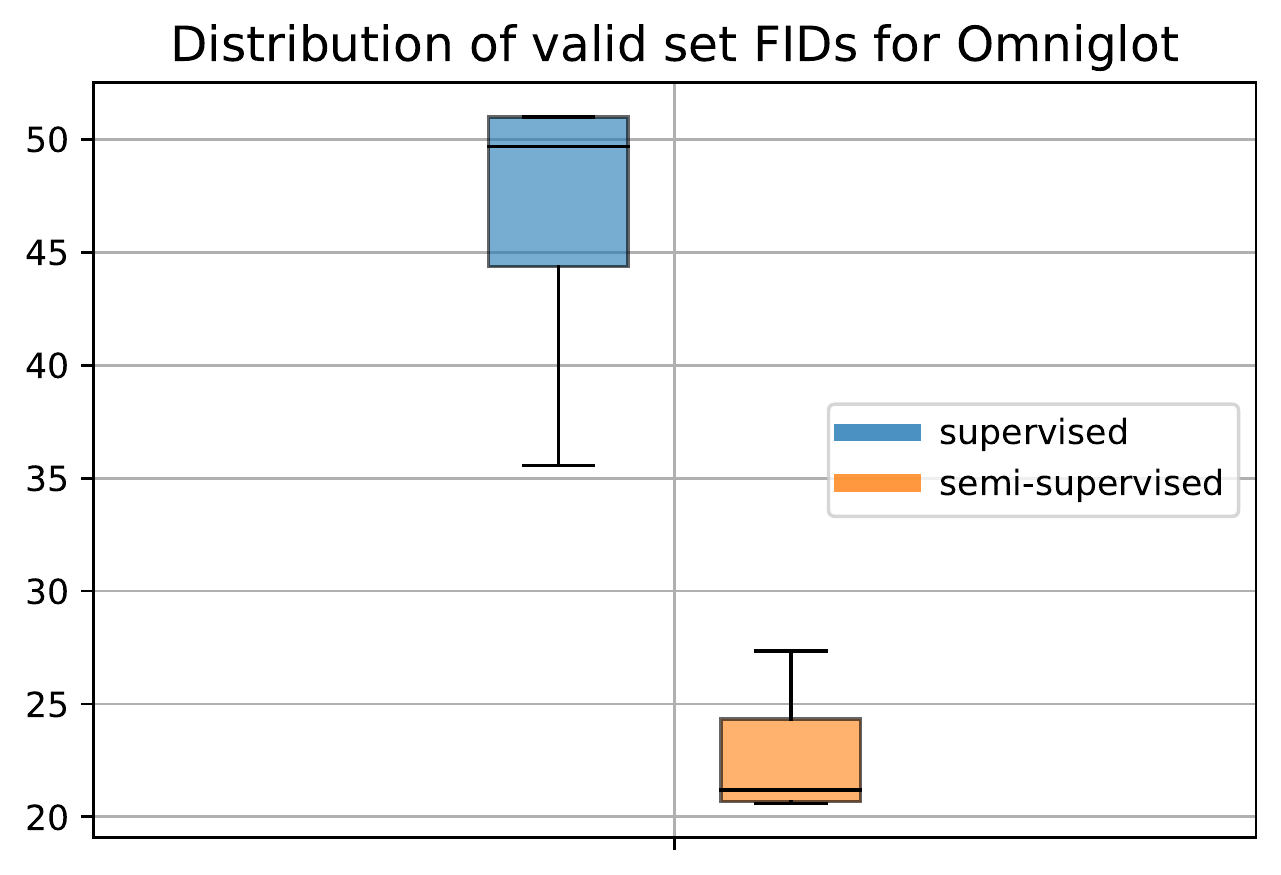}
    \caption{Comparison of valid set FIDs between the supervised and semi-supervised formulations for Omniglot.}
    \label{fig:boxplot_og}
\end{figure}

\newpage

\subsection{Additional results}

\subsubsection{CIFAR-100}

\begin{figure}[h]
	\centering
	\begin{subfigure}[b]{0.4\textwidth}
		\centering
		\includegraphics[width=\textwidth]{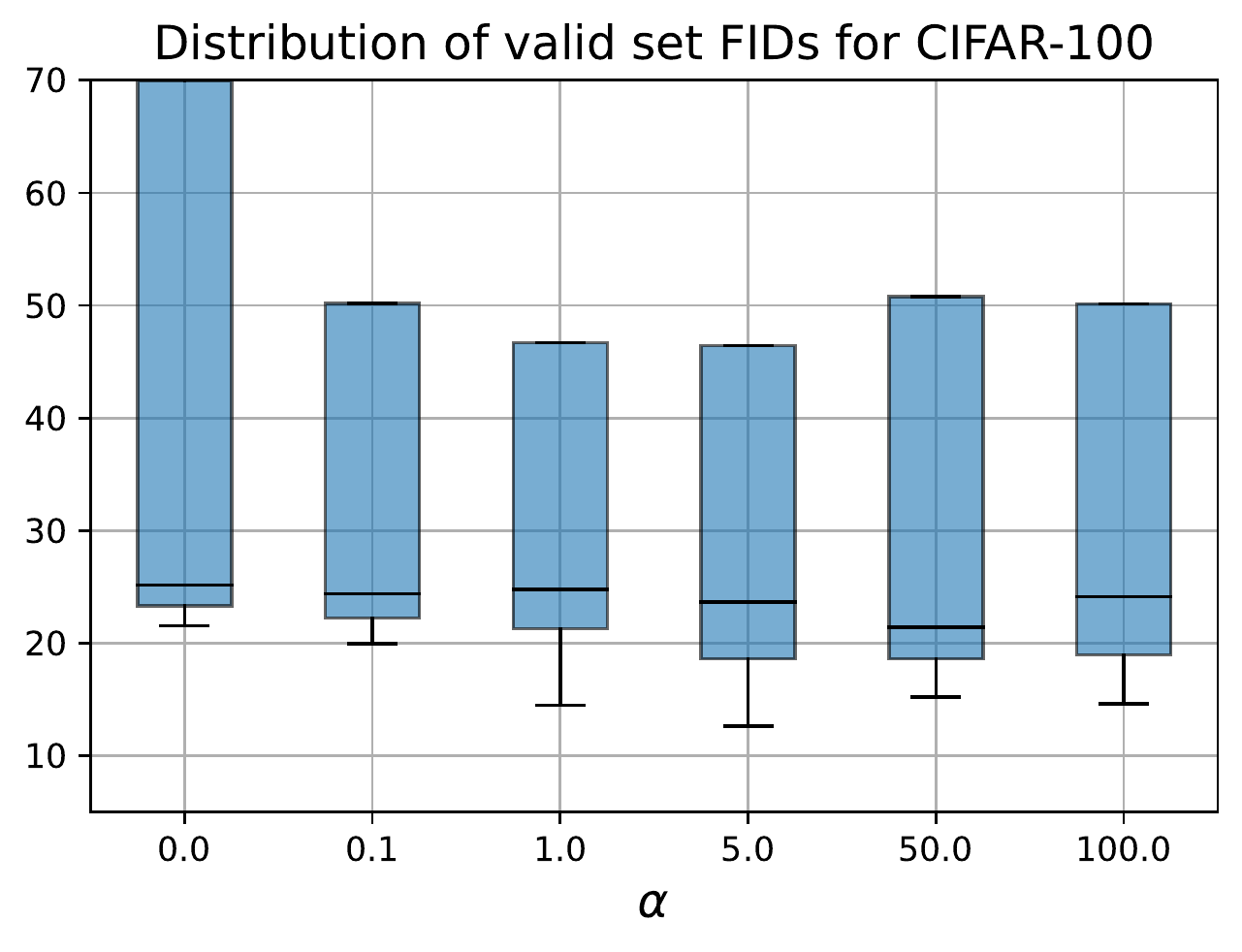}
		\caption{$k=5$}
		\label{fig:c100_fids_alpha}
	\end{subfigure}
	\begin{subfigure}[b]{0.4\textwidth}
		\centering
		\includegraphics[width=\textwidth]{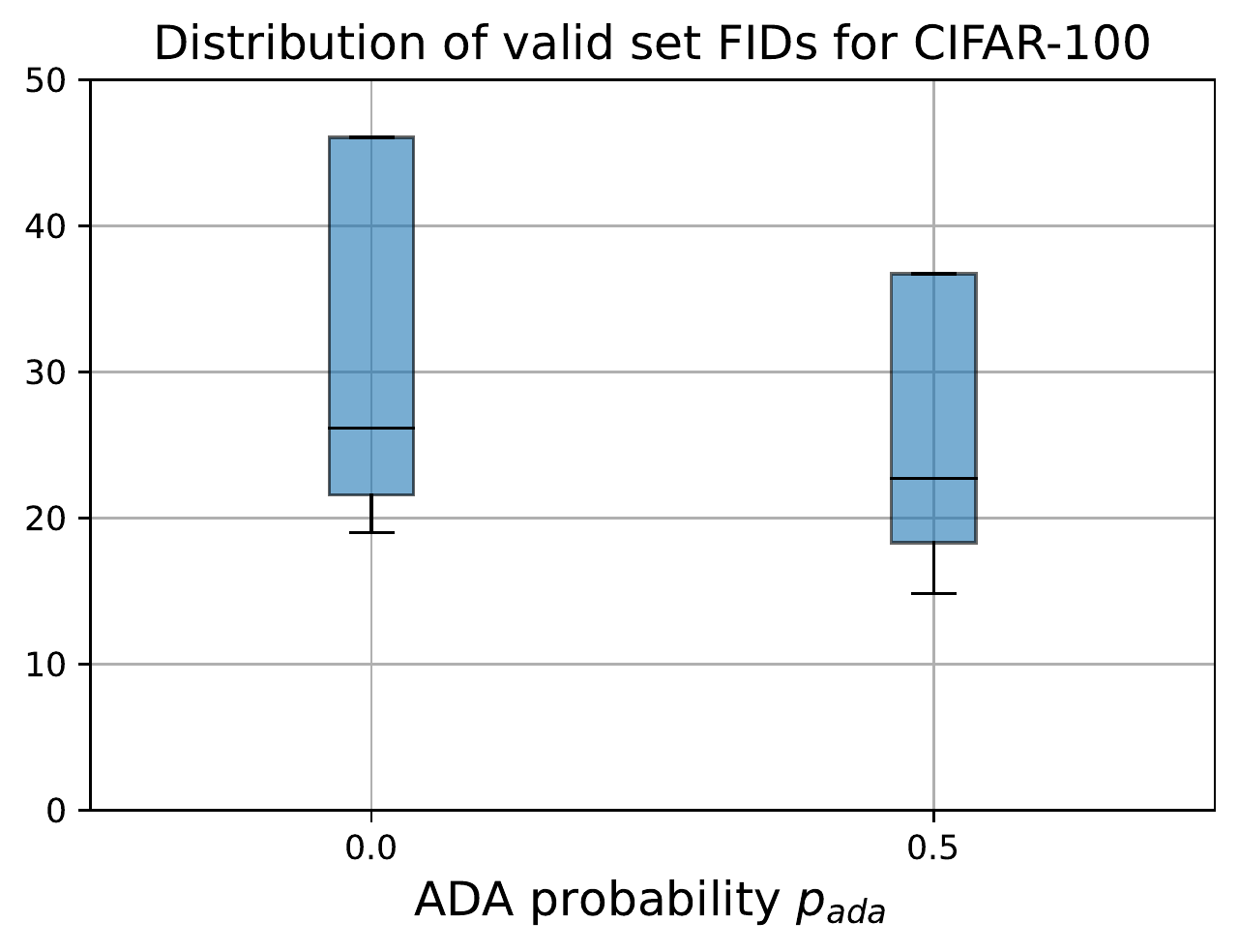}
		\caption{$k=15$}
		\label{fig:c100_fids_p}
	\end{subfigure}
	\caption{For \ref{fig:c100_fids_alpha}, the distributions are computed over all possible enumerations of $\dfm \in \text{\{linear, all\}}$, $\gfm \in \text{\{linear, all\}}$ and $p_{\text{ada}} \in \{0.0, 0.5\}$. For \ref{fig:c100_fids_p}, the distributions are computed over all possible enumerations of $\dfm \in \text{\{linear, all\}}$, $\gfm \in \text{\{linear, all\}}$ and $\alpha \in \{0.0, 0.1, 1.0, 5.0, 50.0, 100.0\}$. }
	\label{fig:c100_fids}
\end{figure}

We further validate our approach on the CIFAR-100 dataset. CIFAR-100 consists of 100 classes of natural images each comprising of 600 images per class. Depending on the dataset seed, we randomly divide the classes into 80\% for the source (training) classes and the remaining 20\% for the target (testing) classes. Furthermore, we use the official BigGAN \citep{biggan} architecture\footnote{\url{https://github.com/ajbrock/BigGAN-PyTorch}} but modified to be more reminiscent of StyleGAN. Concretely, this involves implementing adaptive batch norm for $\bm{z}$, replacing $\bm{z}$-conditioned MLP at the start of the generator with a learnable tensor, implementing exponential moving averages for the weights of the generator \citep{yaz2018unusual} and also adding a minibatch discrimination layer after the convolution layers in the discriminator network. In addition to this, we also experiment with the proposed adaptive discriminator augmentation (ADA) trick proposed in \cite{styleganv2_ada} where data augmentation operations stochastically are applied to both the real and generated images without the data augmentation-specific operations `contaminating' the generative distribution. This is of interest here since such a technique was motivated to be used to improve GANs on limited data. For ADA, we do not use the adaptive variant of the algorithm and simply apply transforms with some probability $p_{\text{ada}}$, where transformations can either be horizontal flips, random rotations, and random translations. In Figure \ref{fig:c100_fids} we show the distribution of FIDs with respect to the semi-supervised coefficient $\alpha$. It can be seen that the best FID is achieved with $\alpha = 5$ (Figure \ref{fig:c100_fids_alpha}). Using discriminator augmentations is also beneficial to decreasing FID (Figure \ref{fig:c100_fids_p}).

\begin{figure}[h]
	\centering
    \includegraphics[width=0.4\textwidth]{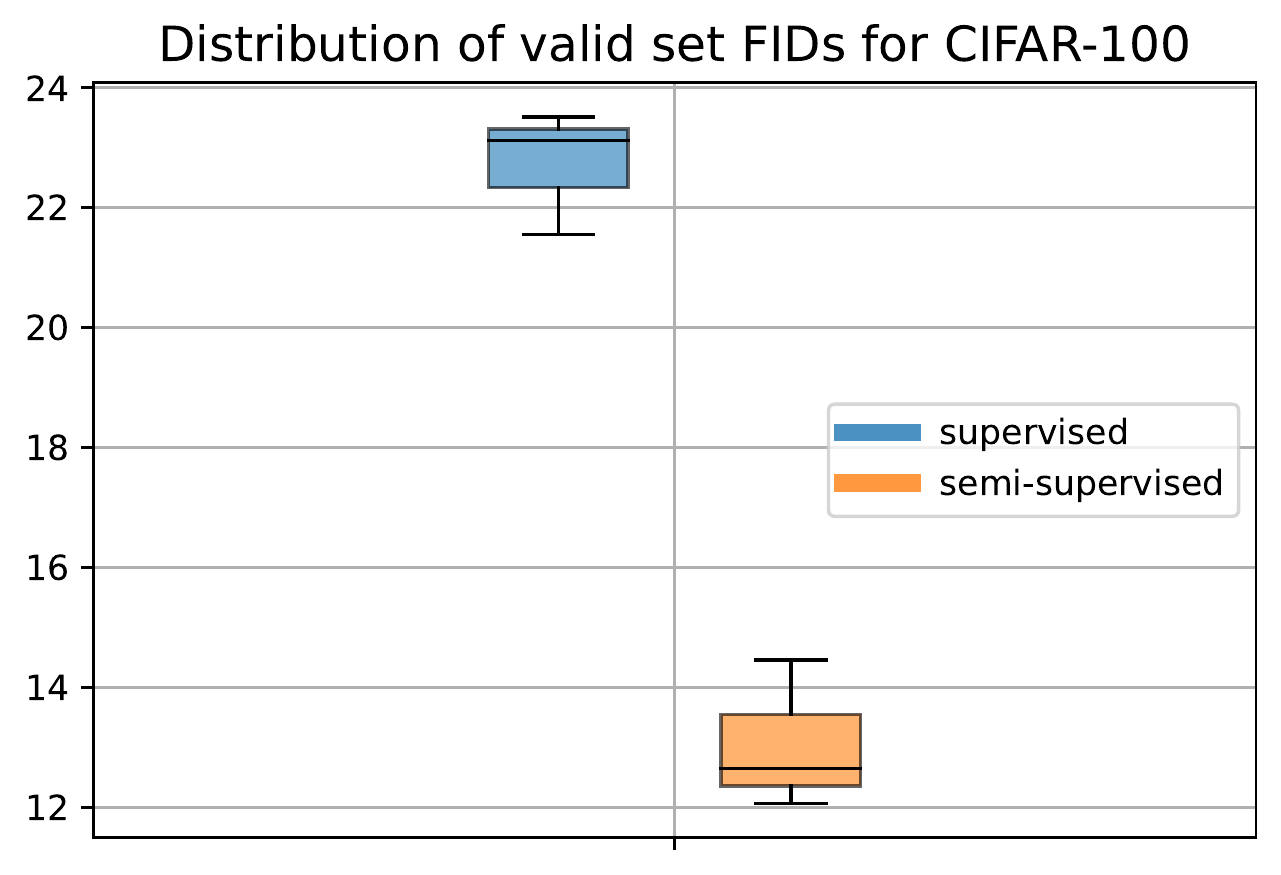}
    \caption{Comparison of valid set FIDs between the supervised and semi-supervised formulations for CIFAR-100.}
    \label{fig:boxplot_c100}
\end{figure}

\begin{table*}[h]
    \caption{Results for CIFAR100, for $k=50$. Results are averages over three different dataset seeds.}
    \centering
    \begin{adjustbox}{max width=0.6\textwidth}
    \begin{tabular}{P{0.01cm} lcc}
      \toprule[0.8pt]
      & Method & $\dvalid$ & $\dtest$ \cr 
      \midrule
      \multirow{3}{*}{\rotatebox[origin=c]{90}{ours}} & Baseline w/ data aug & 73.55 $\pm$ 1.58 & 70.78 $\pm$ 2.65 \cr
      & Fine-tuned GAN & 73.46 $\pm$ 1.46 & 70.79 $\pm$ 3.08 \cr
      & + semi-supervised (Sec. \ref{sec:further_discussion}) & 73.56 $\pm$ 1.56 & \textbf{71.06 $\pm$ 2.31} \cr      
      \bottomrule[0.8pt]
    \end{tabular}
    \end{adjustbox}
    \vspace{0.1mm}
    \label{tb:results_c100}
\end{table*}

\newpage

\subsubsection{EMNIST}

See Table \ref{tb:results_k25}.

\begin{table*}[h]
    \caption{EMNIST results for $k=25$ on both $\dvalid$ and $\dtest$. For each experiment and its dataset seed, the hyperparameters used for the test set evaluation correspond to the optimal hyperparameters found for the same corresponding experiment / seed on the validation set (e.g. \# of epochs, $\alpha$ for input mixup, $n_s$ for GAN).}
    \centering
    \begin{adjustbox}{max width=0.8\textwidth}
    \begin{tabular}{P{0.01cm} lcc}
      \toprule[0.8pt]
      & Method & $\dvalid$ & $\dtest$ \cr 
      \midrule
      \multirow{3}{*}{\rotatebox[origin=c]{90}{ours}} & Baseline w/ data aug & 93.88 $\pm$ 3.75 & 93.39 $\pm$ 3.99 \cr
      & Fine-tuned GAN & 93.85 $\pm$ 3.61 & 93.29 $\pm$ 3.86 \cr
      & + semi-supervised (Sec. \ref{sec:further_discussion}) & \textbf{94.10 $\pm$ 3.53} & \textbf{93.62 $\pm$ 3.80} \cr
      \bottomrule[0.8pt]
    \end{tabular}
    \end{adjustbox}
    \vspace{0.1mm}
    \label{tb:results_k25}
\end{table*}

\subsection{Optimal hyperparameters for GAN fine-tuning}

In order of dataset seeds 0-4:

\subsubsection{EMNIST k=5}

\begin{itemize}
    \item \dfm: ['all', 'embed', 'linear', 'all', 'all']
    \item \gfm: ['embed', 'embed', 'embed', 'embed', 'embed']
    \item $\gamma$: [100.0, 100.0, 100.0, 100.0, 100.0]
\end{itemize}

\subsubsection{EMNIST k=10}

\begin{itemize}
    \item \dfm: ['all', 'all', 'all', 'all', 'all']
    \item \gfm: ['embed', 'linear', 'linear', 'embed', 'embed']
    \item $\gamma$: [100.0, 100.0, 100.0, 100.0, 100.0]
\end{itemize}

\subsubsection{EMNIST k=15}

\begin{itemize}
    \item \dfm: ['all', 'linear', 'all', 'linear', 'all']
    \item \gfm: ['embed', 'embed', 'linear', 'embed', 'embed']
    \item $\gamma$: [100.0, 100.0, 100.0, 100.0, 100.0]
\end{itemize}

\subsection{Optimal hyperparameters for GAN fine-tuning (semi-supervised)}

In order of dataset seeds 0-4:

\subsubsection{Omniglot k=5}

\begin{itemize}
    \item \dfm: ['all', 'all', 'all', 'all']
    \item \gfm: ['linear', 'linear', 'linear', 'linear']
    \item $\gamma$: [10.0, 10.0, 20.0, 20.0],
    \item max epochs trained for: ['215 $\times$ 100', '645 $\times$ 100', '520 $\times$ 100', '354 $\times$ 100']
\end{itemize}

\subsection{Optimal hyperparameters for classifier fine-tuning} \label{sec:appx_hps}

In the order of dataset seeds 0-4:

\subsubsection{EMNIST k=5}

\begin{itemize}
    \item \textbf{Baseline}: minScale = [0.6, 0.2, 0.8, 0.6, 0.8], 0.60 $\pm$ 0.22
    \item \textbf{Input mixup}: $\alpha$ = [0.1, 0.1, 0.1, 0.1, 0.1], 0.10 $\pm$ 0.00
    \item \textbf{Fine-tuned GAN}: $n_s$ = [15, 5, 5, 10, 20], 11.00 $\pm$ 5.83 
    \item \textbf{Semi-supervised}: $n_s$ = [50, 10, 50, 20, 10], 28.00 $\pm$ 18.33
\end{itemize}

\subsubsection{EMNIST k=10}

\begin{itemize}
    \item \textbf{Baseline}: minScale = [0.8, 0.6, 0.8, 0.4, 0.6], 0.64 $\pm$ 0.15
    \item \textbf{Input mixup}: $\alpha$ = [0.1, 0.1, 0.1, 0.1, 0.1], 0.10 $\pm$ 0.00
    \item \textbf{Fine-tuned GAN}: $n_s$ = [5, 10, 10, 20, 15], 12.00 $\pm$ 5.10 
    \item \textbf{Semi-supervised}: $n_s$ = [10, 10, 20, 50, 50], 28.00 $\pm$ 18.33
\end{itemize}

\subsubsection{EMNIST k=15}

\begin{itemize}
    \item \textbf{Baseline}: minScale = [0.8, 0.4, 0.8, 0.8, 0.4], 0.64 $\pm$ 0.20
    \item \textbf{Input mixup}: $\alpha$ = [0.1, 0.1, 0.1, 0.1, 0.1], 0.10 $\pm$ 0.00
    \item \textbf{Fine-tuned GAN}: $n_s$ = [5, 5, 10, 10, 20], 10.00 $\pm$ 5.48 
    \item \textbf{Semi-supervised}: $n_s$ = [20, 20, 20, 20, 50], 26.00 $\pm$ 12.00
\end{itemize}

\subsubsection{Omniglot k=5}

\begin{itemize}
    \item \textbf{Baseline}: minScale = [0.8, 0.8, 1.0, 1.0, 0.4]
    \item \textbf{Input mixup}: $\alpha$ = [0.1, 0.1, 0.1, 0.1, 0.1]
    \item \textbf{Fine-tuned GAN}: $n_s$ = [1, 1, 1, 3, 2] 
    \item \textbf{Semi-supervised}: $n_s$ = [1, 1, 1, 1, 2]
\end{itemize}

\newpage

\subsection{Generated images for EMNIST (novel classes)}

\begin{figure}[h!]
	\centering
	\begin{subfigure}[b]{0.6\textwidth}
		\centering
		\includegraphics[width=\textwidth]{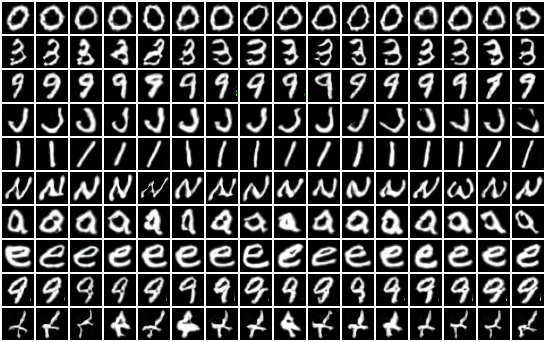}
		\caption{$k=5$}
		\label{fig:appx_generated_images_k5}
	\end{subfigure} \\ 
	\begin{subfigure}[b]{0.6\textwidth}
		\centering
		\includegraphics[width=\textwidth]{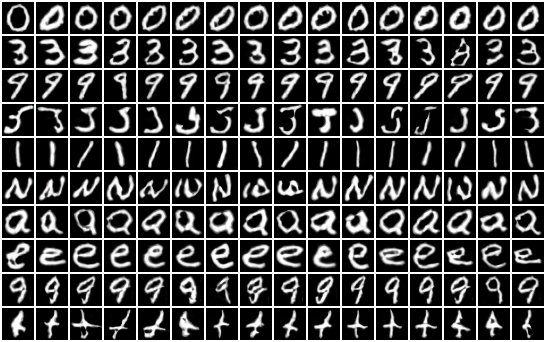}
		\caption{$k=10$}
		\label{fig:appx_generated_images_k10}
	\end{subfigure} \\ 
	\begin{subfigure}[b]{0.6\textwidth}
		\centering
		\includegraphics[width=\textwidth]{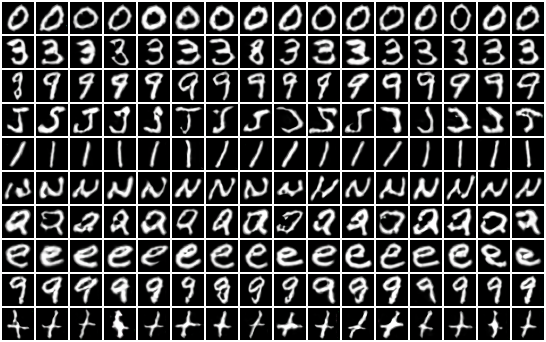}
		\caption{$k=15$}
		\label{fig:appx_generated_images_k15}
	\end{subfigure}
	\caption{Generated samples from fine-tuned GAN in Section \ref{sec:gan_finetuning}.}
	\label{fig:appx_generated_images}
\end{figure}

\newpage

\subsection{Generated images for EMNIST (novel classes, semi-supervised)}

\begin{figure}[h!]
	\centering
	\begin{subfigure}[b]{0.6\textwidth}
		\centering
		\includegraphics[width=\textwidth]{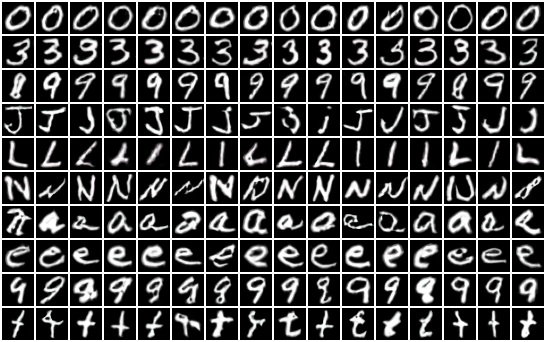}
		\caption{$k=5$}
		\label{fig:appx_generated_images_semi_k5}
	\end{subfigure} \\ 
	\begin{subfigure}[b]{0.6\textwidth}
		\centering
		\includegraphics[width=\textwidth]{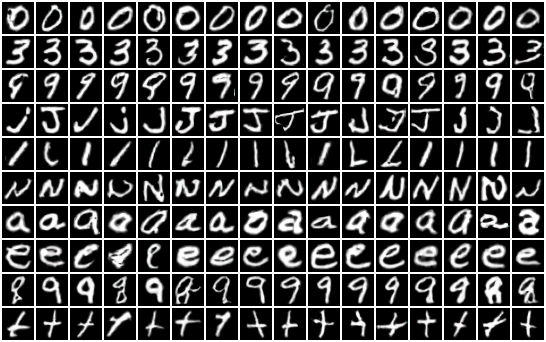}
		\caption{$k=10$}
		\label{fig:appx_generated_images_semi_k10}
	\end{subfigure} \\ 
	\begin{subfigure}[b]{0.6\textwidth}
		\centering
		\includegraphics[width=\textwidth]{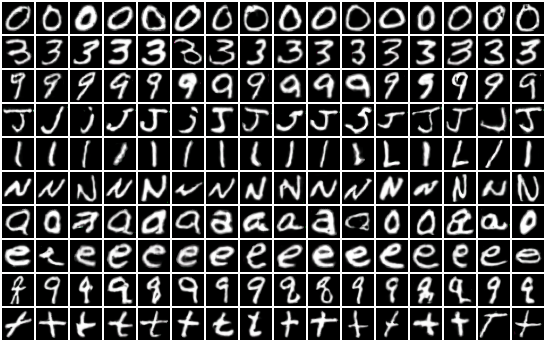}
		\caption{$k=15$}
		\label{fig:appx_generated_images_semi_k15}
	\end{subfigure}
	\caption{Generated samples using the semi-supervised formulation in Section \ref{sec:further_discussion}.}
	\label{fig:appx_generated_images_semi}
\end{figure}

\newpage

\subsection{Generated images for Omniglot (novel classes)}

\begin{figure}[h]
    \centering
    \includegraphics[trim={0px 0px 0px 6285px},clip, width=0.7\textwidth]{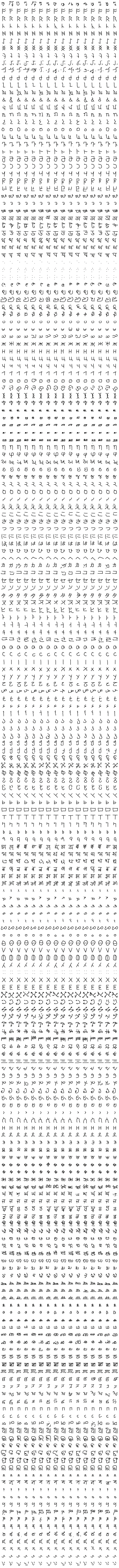}
    \caption{$k=5$. Only a subsample of classes are shown here due to space constraints.}
    \label{fig:appx_generated_images_k5_og}
\end{figure}

\newpage

\subsection{Generated images for Omniglot (novel classes, semi-supervised)}

\begin{figure}[h]
    \centering
    \includegraphics[trim={0px 0px 0px 6285px},clip, width=0.7\textwidth]{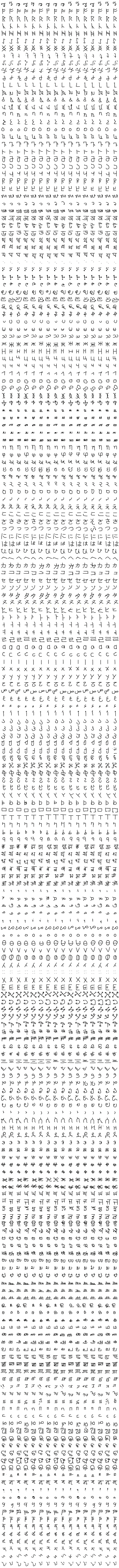}
    \caption{$k=5$. Only a subsample of classes are shown here due to space constraints.}
    \label{fig:appx_generated_images_k5_semi_og}
\end{figure}

\newpage

\subsection{Generated images for CIFAR100 (novel classes)}

\begin{figure}[h]
    \centering
    \includegraphics[width=0.7\textwidth]{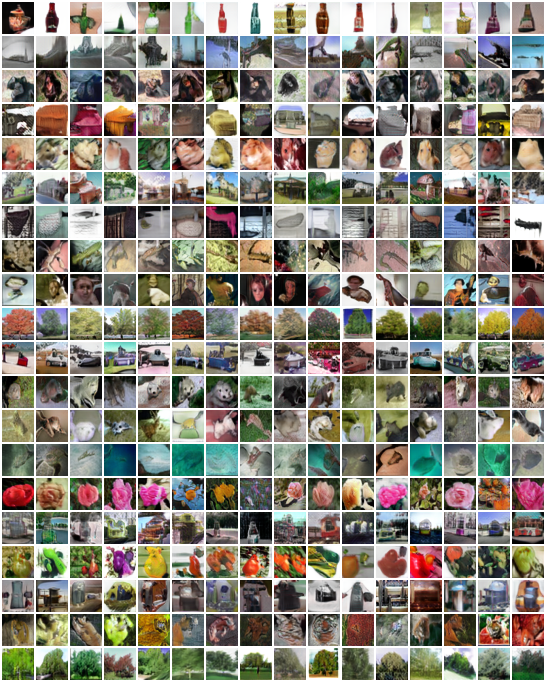}
    \caption{$k=50$.}
    \label{fig:appx_generated_images_k50_c100}
\end{figure}

\newpage

\subsection{Generated images for CIFAR100 (novel classes, semi-supervised)}

\begin{figure}[h]
    \centering
    \includegraphics[width=0.7\textwidth]{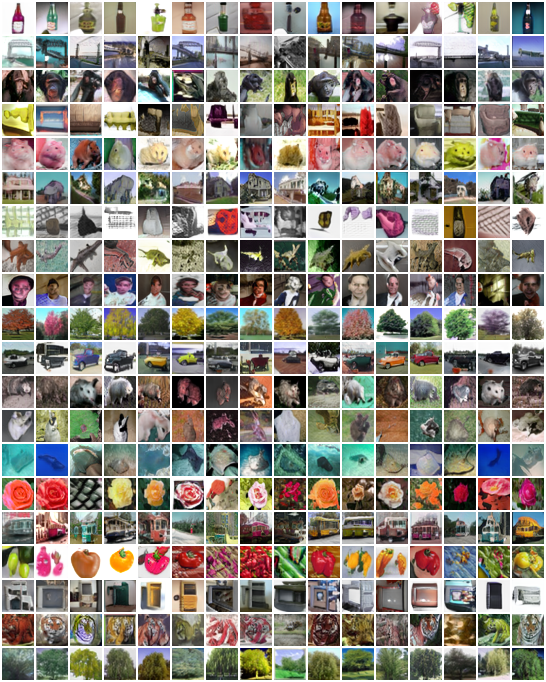}
    \caption{$k=50$.}
    \label{fig:appx_generated_images_k50_semi_c100}
\end{figure}

\newpage

\subsection{Dataset splits}

\begin{figure}[h!]
	\centering
	\includegraphics[width=0.75\textwidth]{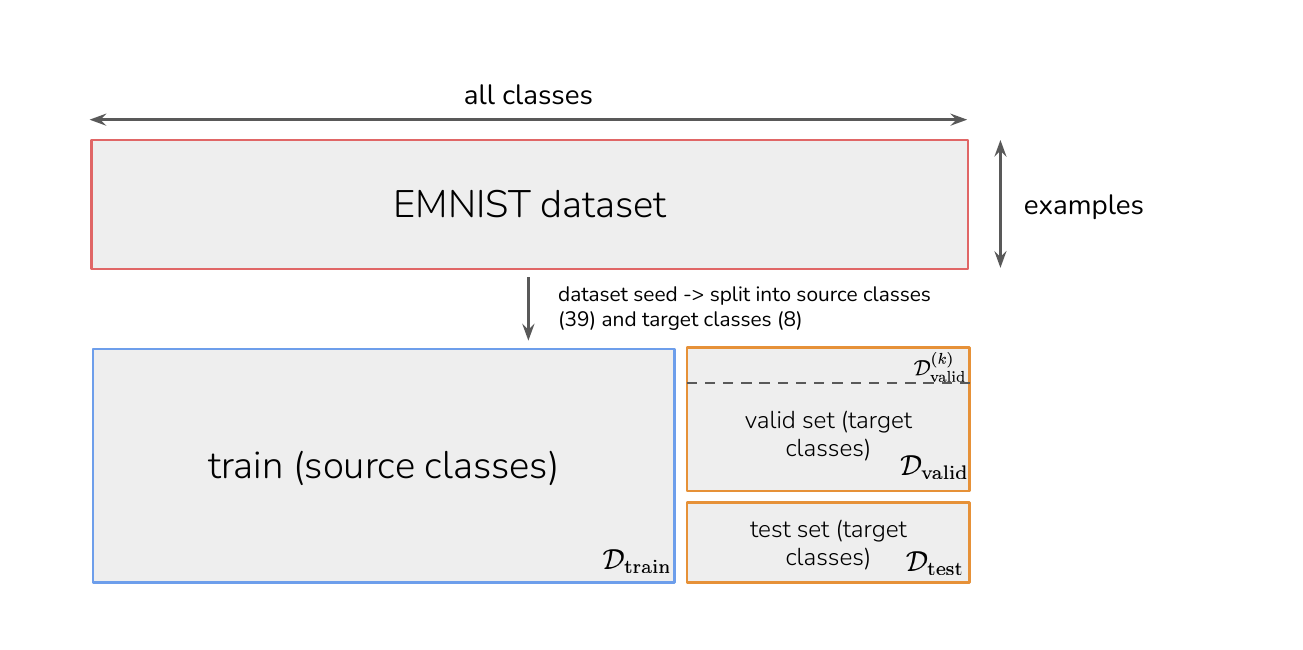}
	\caption{Illustration of how we split our dataset. The dataset seed is used to control what classes comprise training (source) classes and validation/test (target) classes. The GAN and classifier are trained on $\dtrain$. The GAN is then fine-tuned on $\dsupp$ and is used to generate examples for the target classes. The classifier pretrained on $\dtrain$ is now fine-tuned on $\dsupp$ and the GAN-generated examples and the goal is to maximise performance on $\dvalid$, which is done through hyperparameter tuning. When this is completed, classifier performance is evaluated on the held-out test set $\dtest$.}
	\label{fig:appendix_data_splits}
\end{figure}

\begin{figure}[h!]
	\centering
	\includegraphics[width=0.75\textwidth]{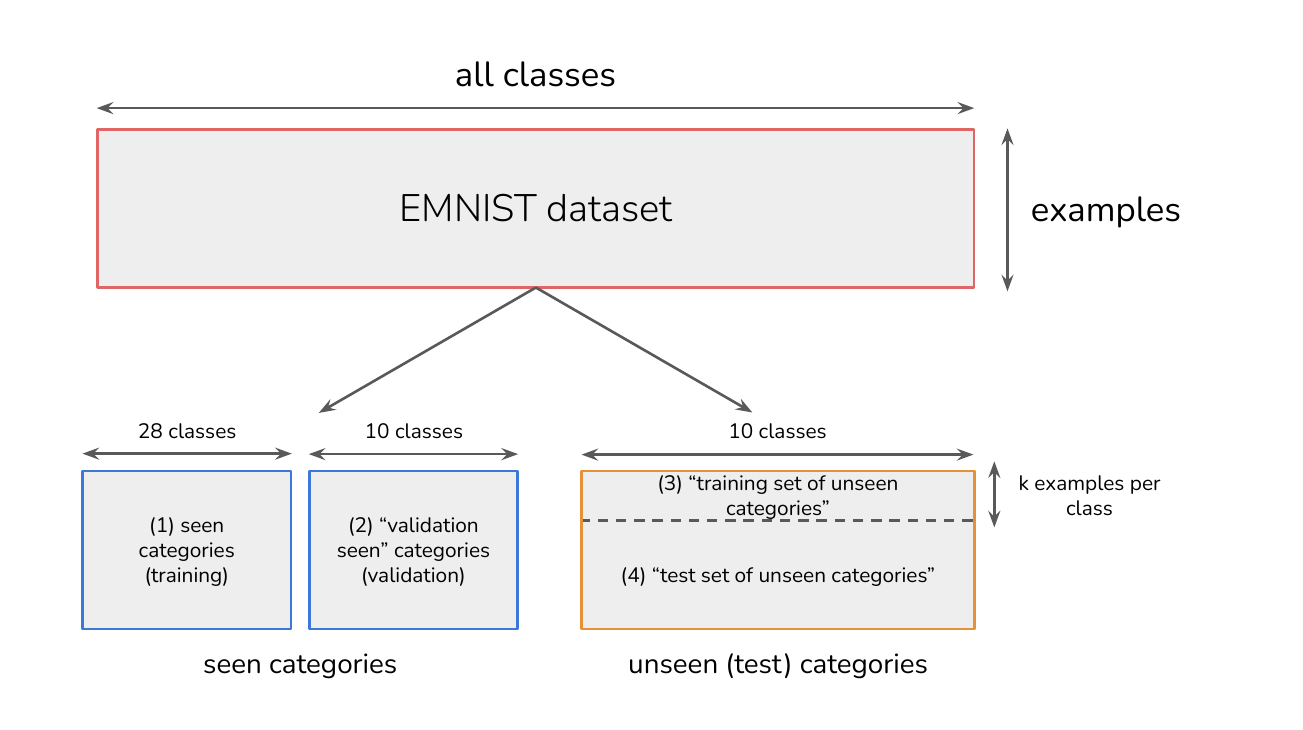}
	\caption{Illustration of how the dataset is split according to MatchingGAN \citep{hong2020matchinggan}. The training set (1) is used to train the GAN, and validation set (2) is used to monitor GAN training (tune GAN training hyperparameters via FID?). What we call the `support set' they have called the `training set of unseen categories' (3), and this is what the GAN uses to generate additional examples. The classifier pre-trained on (1) is now fine-tuned on (3) + the GAN-generated examples and is used to predict accuracy on (4). As we mentioned in Section \ref{sec:exps_and_results}, hyperparameter tuning of the generation process / classifier can lead to biased estimates of performance on (4), since there is no additional held-out test set. Furthermore, in F2GAN \citep{f2gan} no validation set appears to be mentioned.}
	\label{fig:appendix_matchinggan_setup}
\end{figure}

\begin{figure}[t]
	\centering
	\includegraphics[width=0.75\textwidth]{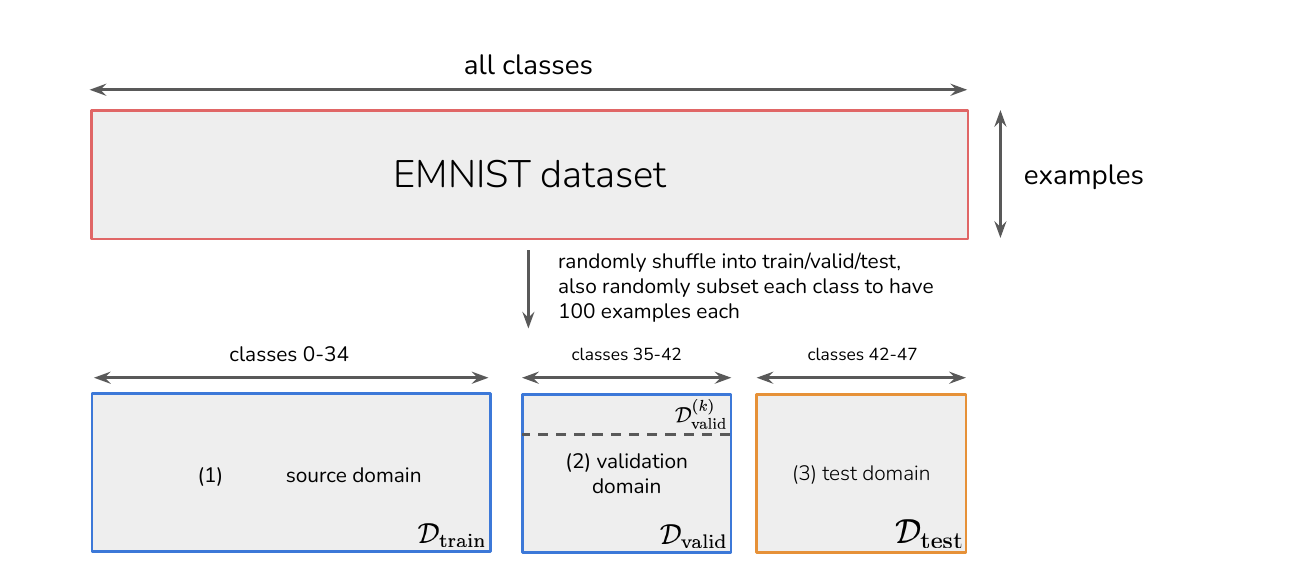}
	\caption{Illustration of how the dataset is split according to DAGAN \citep{dagan}. Some details regarding the evaluation were not clear, but from what can be gathered, DAGAN is trained on the source domain (training set). A baseline classifier is also trained (from scratch, not fine-tuned) on the validation domain, and hyperparameter tuning is done for $n_s$ to determine how many DAGAN-generated samples per class are optimal. The final evaluation is then training a classifier from scratch on the target domain with the optimal $n_s$ found via hyperparameter tuning. The test accuracy is averaged over five independent runs (dataset seeds).}
	\label{fig:appendix_dagan_setup}
\end{figure}

\newpage

\subsection{Semi-supervised experiments}

\begin{figure}[h]
	\centering
	\includegraphics[width=0.75\textwidth]{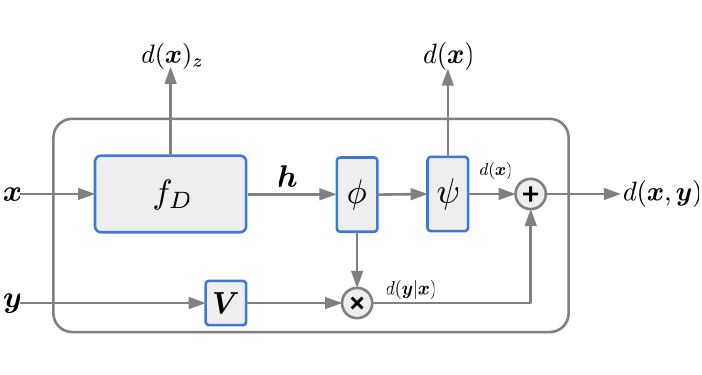}
	\caption{Projection discriminator from \cite{cgan_proj}. Recall from Section \ref{sec:further_discussion} that the output of the discriminator is $d(\bm{x}, \bm{y}) = d(\bm{y}|\bm{x}) + d(\bm{x}) = \bm{y}^{T}\bm{V} \cdot \phi(f_D(\bm{x})) + \psi(\phi(f_D(\bm{x})))$. The fact that $d(\bm{x}, \bm{y})$ partly decomposes into $d(\bm{x})$ means that the projection discriminator can be leveraged in a semi-supervised manner.}
	\label{fig:appendix_cgan}
\end{figure}

\end{document}